\documentclass[sigconf]{acmart}
\usepackage{amsmath,amsthm, mathrsfs} 
\usepackage{dsfont}
\usepackage{graphicx}
\usepackage{float}
\usepackage{color}
\usepackage[linesnumbered,lined,boxed,commentsnumbered]{algorithm2e}
\usepackage{subcaption}
\usepackage{enumitem}
\usepackage{xcolor}
\usepackage{tabularx}
\usepackage{balance}
\AtBeginDocument{%
  \providecommand\BibTeX{{%
    \normalfont B\kern-0.5em{\scshape i\kern-0.25em b}\kern-0.8em\TeX}}}

\hypersetup{
	colorlinks=true,
	urlcolor=magenta,
}

\copyrightyear{2023}
\acmYear{2023}
\setcopyright{acmlicensed}\acmConference[KDD '23]{Proceedings of the 29th ACM SIGKDD Conference on Knowledge Discovery and Data Mining}{August 6--10, 2023}{Long Beach, CA, USA}
\acmBooktitle{Proceedings of the 29th ACM SIGKDD Conference on Knowledge Discovery and Data Mining (KDD '23), August 6--10, 2023, Long Beach, CA, USA}
\acmPrice{15.00}
\acmDOI{10.1145/3580305.3599254}
\acmISBN{979-8-4007-0103-0/23/08}

\begin{document}

\title{Adversarial Constrained Bidding via Minimax Regret Optimization with Causality-Aware Reinforcement Learning}
% \title{Causality-Aware Minimax Regret Optimization for Constrained Bidding in Black-box Adversarial Environments}

% \title{Constrained Bidding in Adversarial Environments \\
% through Minimax Regret Optimization}

%%%%%%%%%%%%%%%%% under review, anonymous
\author{Haozhe Wang}
\authornote{Correspondence to Haozhe Wang <jasper.whz@outlook.com>}
% \email{jasper.whz@outlook.com}
\affiliation{%
  \institution{Alibaba Group}
  % \streetaddress{P.O. Box 1212}
   \city{Beijing}
  \country{China}
}
\author{Chao Du}
% \email{jasper.whz@outlook.com}
% \email{duchao0726@gmail.com}
\affiliation{%
  \institution{Alibaba Group}
  % \streetaddress{P.O. Box 1212}
  \city{Beijing}
  \country{China}
}

\author{Panyan Pang}
% \author{Li He}
% \author{Liang Wang}
% \author{Bo Zheng}
% \email{jasper.whz@outlook.com}
\affiliation{%
  \institution{Alibaba Group}
  % \streetaddress{P.O. Box 1212}
  \city{Beijing}
  \country{China}
}

\author{Li He}
\affiliation{%
  \institution{Alibaba Group}
  % \streetaddress{P.O. Box 1212}
  \city{Beijing}
  \country{China}
}
\author{Liang Wang}
\affiliation{%
  \institution{Alibaba Group}
  % \streetaddress{P.O. Box 1212}
  \city{Beijing}
  \country{China}
}
\author{Bo Zheng}
\affiliation{%
  \institution{Alibaba Group}
  % \streetaddress{P.O. Box 1212}
  \city{Beijing}
  \country{China}
}
\newcommand{\wang}[1]{\textcolor{blue}{#1}} 
\newcommand{\fang}[1]{\textcolor{yellow}{[FL:#1]}} 
\newcommand{\chao}[1]{\textcolor{red}{[CD: #1]}}
\begin{abstract}
  The proliferation of the Internet has led to the emergence of online advertising, driven by the mechanics of online auctions. In these repeated auctions, software agents participate on behalf of aggregated advertisers to optimize for their long-term utility. To fulfill the diverse demands, bidding strategies are employed to optimize advertising objectives subject to different spending constraints. Existing approaches on constrained bidding typically rely on i.i.d. train and test conditions, which contradicts the adversarial nature of online ad markets where different parties possess potentially conflicting objectives. In this regard, we explore the problem of constrained bidding in adversarial bidding environments, which assumes no knowledge about the adversarial factors. Instead of relying on the i.i.d. assumption, our insight is to align the train distribution of environments with the potential test distribution meanwhile minimizing policy regret. Based on this insight, we propose a practical Minimax Regret Optimization (MiRO) approach that interleaves between a teacher finding adversarial environments for tutoring and a learner meta-learning its policy over the given distribution of environments. In addition, we pioneer to incorporate expert demonstrations for learning bidding strategies. Through a causality-aware policy design, we improve upon MiRO by distilling knowledge from the experts. Extensive experiments on both industrial data and synthetic data show that our method, MiRO with Causality-aware reinforcement Learning (MiROCL), outperforms prior methods by over $30\%$.
\end{abstract}

%%%%%%%%%%%%%%%%%%%%%%%%%%%%%%%% CCS TBD

% \ccsdesc[500]{Information systems~Display advertising}
% \ccsdesc[300]{Mathematics of computing~Causal networks}
% \ccsdesc[300]{Computing methodologies~Learning from demonstrations}
%%%%%%%%%%%%%%%%%%%%% CCS TBD



%%%%%%%%%%%%%%%%%%% Not used 

\maketitle
% \vspace{+.1cm}
\vspace{-.25cm}
\section{Introduction}
\def\mediaplot{
\begin{figure}[t]
    \centering
    \includegraphics[width=\linewidth]{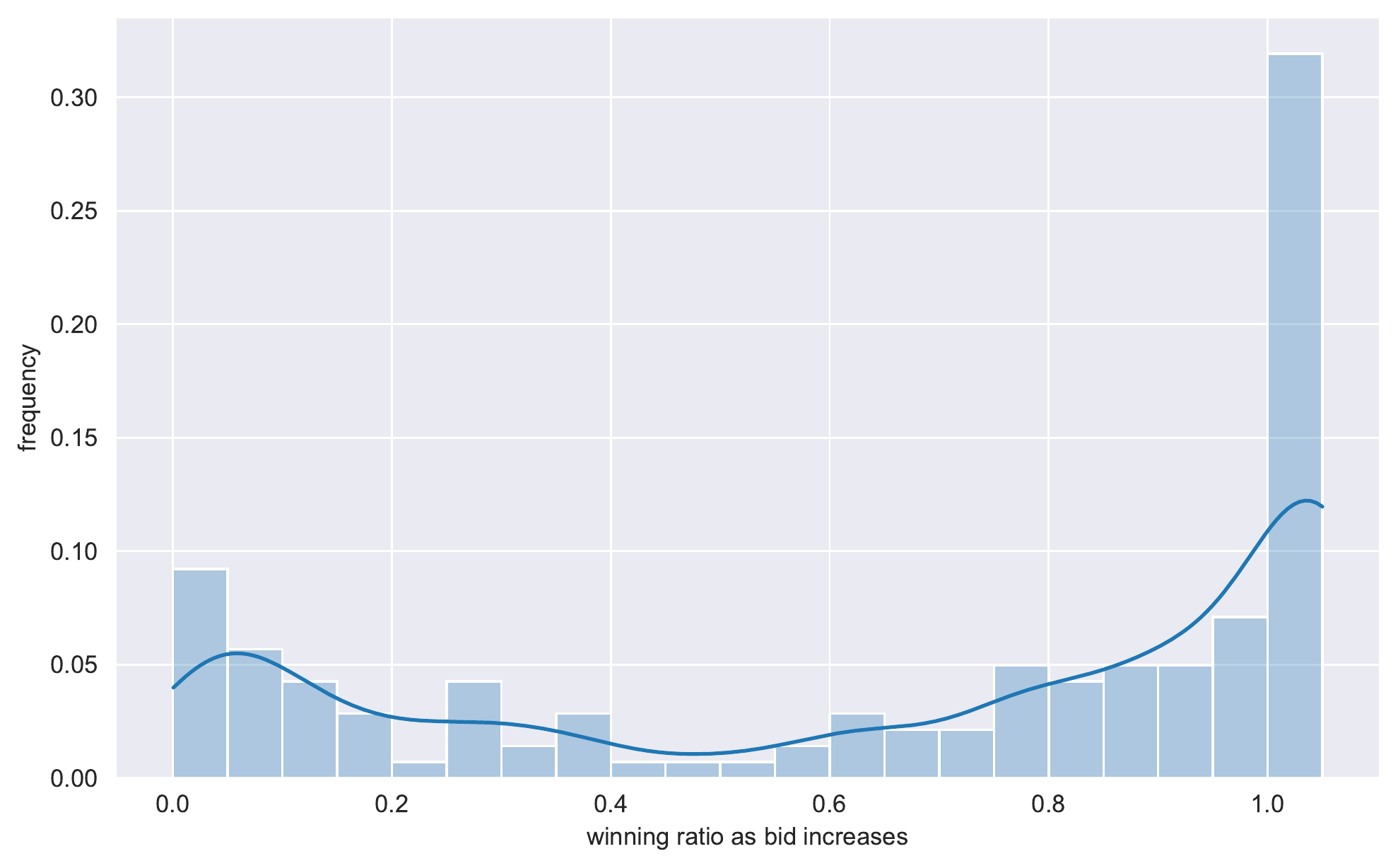}
    \caption{Evidence of real-world adversarial situations. X-axis: the ratio of winning auctions under fixed cost. Y-axis: the normalized count of display positions.  We observe that the winning ratio in a number of dispaly positions (i.e., publishers) becomes closer to $0$, implying increased cost for increased bids when facing some sellers. }
    \label{fig:media}
    \vspace{-.4cm}
\end{figure}
}

The proliferation of the Internet has led to the rise of online advertising as a multi-billion dollar industry. At the heart of online advertising lies online auctions~\cite{defauction}, where publishers repeatedly sell ad slots to advertisers seeking brand promotion, greater conversions, etc. Traditionally, incentive-compatible auctions such as second-price auctions are widely adopted, as they possess the desirable property of `truthful bidding' for myopic bidders -- truthfully revealing private values is optimal for these non-strategic bidders in order to maximize their immediate utility~\cite{ic, repeatauctionsurvey}. 

However, the crucial assumption of myopic bidders has become obsolete in recent times and truthful bidding does not optimize advertisers' long-term utilities. Demand-side platforms (DSPs), as an intermediary on behalf of aggregated advertisers, is now the actual entity participating in billions of auctions a day. Rather than truthful bidding, DSP agents employ bidding strategies to satisfy the demands of the diversified advertisers, who often seek to maximize certain utilities while subject to spending constraints~\cite{wilkens2016mechanism}. For instance, brand advertisers seek long-term growth and awareness and typically optimize for metrics such as impressions, clicks, and subject to return-on-investment (ROI) constraints that require a minimum ratio of utilities to cost. 

In order to cater to the diverse demands of the advertisers, extensive studies have been conducted on designing and learning bidding strategy. The existing literature can be broadly classified based on the setting of constraints. The majority of research has been focused on bidding subject to at most a budget constraint~\cite{budget0, budget1,budget2,budget3,pacing1,mpc1,RLB}, which may not fully capture the diversity of spending constraints in the field. To address this limitation, a few studies~\cite{PID, USCB, CBRL} have explored optimal bidding with ROI-like constraints. The ROI-Constrained Bidding (RCB) problem, which involves ensuring that the ROI-like constraints, such as acquisition-per-cost and click-per-cost, exceed a prescribed limit while adhering to a budget constraint, is viewed as a prototypical problem that generalize to the diverse advertising objectives~\cite{CBRL}.

% In addition, the bidder's private value can also vary over time due to changes in user intent driven by exogenous factors, thus affecting the utility estimation of its ads~\cite{onlearn4}. 

Despite promising results obtained by previous methods~\cite{USCB,CBRL}, they typically follows the Empirical Risk Minimization (ERM) principle~\cite{batch1,CASTR} that relies on the assumption of independent and identically distributed (i.i.d.) train and test conditions. This contradicts many real-world situations where the sellers and other rival bidders behave adversarially, as all parties seek to optimize their own utilities that are potentially conflicting with one another~\cite{repeatauctionsurvey}. For instance, sellers may learn the distribution of bidders' private values and set personalized reserved prices in auctions~\cite{yahoo, reserve1, reserve2}. Recent studies~\cite{regretnet, auctionlearn} have introduced neural network-based selling mechanisms learned from data. In addition, rival bidders can also employ complex bidding strategies to optimize their long-term utility~\cite{dual}, leading to a complex distribution of competing bids that can affect the performance of our bidding agent~\cite{advbid1}. These considerations point out the inherently adversarial nature of the bidding environment (also summarized as non-stationarity in \cite{CBRL}).

The problem of bidding in adversarial environments has remained largely unexplored, with only a few recent works~\cite{advbid1, advbid2} showing developments. These works focused on adversarial bidding \emph{in the absence of} constraints in \emph{first-price auctions}~\cite{advbid1} or relied on \emph{assumptions about the adversaries}~\cite{advbid2}. However, there remains an uncharted problem that we aim to investigate in this paper, namely constrained bidding in black-box adversarial environments, which assumes no knowledge about how exogenous factors cause adversarial perturbations to the bidding environment. From a game-theoretic standpoint, a black-box adversary purposefully perturb the bidding environment, such as altering the market dynamics or the value distribution, exploiting its knowledge about the bidder. The bidder is therefore subjected to test environments where it potentially perform worse, and it must behave adaptively in the adversarial environments to achieve optimal performance.

To address the issue of broken i.i.d. assumption, our basic insight is to align the train distribution of environments with the potential test distribution, meanwhile minimizing policy regret, i.e., the performance discrepancy between the policy and a pre-computed optimal policy (referred to as the `expert' policy). Based on this insight, we derive a Minimax Regret Optimization (MiRO) framework which interleaves between identifying the aligning train distribution in the inner supremum problem and optimizing the regret-minimizing policy under such distribution in the outer infimum problem. While MiRO appears appealing, we discovered that both the inner and the outer problems have practical limitations that necessitate refinement to achieve optimal performance:
\vspace{-.1cm}
\begin{itemize}[leftmargin=*]
    \item The inner problem (Sec.~3.2). For lack of knowledge about the exact function structure of the environments and the adversarial factors, it is infeasible to obtain a tractable inner supremum problem that directly optimizes for a distribution. To address this issue, we propose a data-driven approach that learns the latent representation of the adversarial factors by reconstructing the causal structure of world model. This renders a differentiable game that can be optimized end-to-end. 
    % we propose to learn to project the environments into a continuous embedding space (Sec.3.2.1), and make the black-box differentiable by learning a reward estimator (Sec.3.2.2). 
    % Accordingly, we are able to optimize for a train distribution of environments and improve policy under this train distribution both in an end-to-end fashion.
    \item The outer problem (Sec.~3.3). Although regret minimization aims to close the gap between  the policy and the expert, policy learning indeed degenerates to a value maximization problem with the experts taking no effects. To circumvent this issue, we seek to explicitly utilize the useful knowledge from the experts to guide policy learning. Surprisingly, we find that the straight-forward behavioral cloning approach does not work due to the unobserved confounding issue~\cite{causality}. To overcome this challenge, we develop a causality-aware alignment strategy that factorizes into a sub-policy mimicking the causal structure of the experts. 
   
\end{itemize}

The effectiveness and the generalizability of our method, Minimax Regret Optimization with Causality-aware reinforcement Learning (MiROCL), is validated through both large-scale synthetic data and industrial data.

\section{Background and Preliminaries}\label{sec:background}

\def\E{\mathbb{E}}
\def\lb{\left[}
\def\rb{\right]}
\def\S{\mathcal{S}}
\def\O{\mathcal{O}}
\def\A{\mathcal{A}}
\def\M{\mathcal{M}}
\def\reg{\text{Reg}}
\def\1{\mathds{1}}
\def\T{\mathcal{T}}
\def\W{\mathcal{W}}
\def\H{\mathcal{H}}
\def\roi{\text{ROI}}
\def\MO{\M_{\boldsymbol{\omega}}}
\def\boldo{{\boldsymbol\omega}}
\def\U{\mathcal{U}}
\def\D{\mathcal{D}}
\def\J{\mathcal{J}}
\newcommand{\defeq}{\mathrel{\overset{\makebox[0pt]{\mbox{\normalfont\tiny\sffamily def}}}{=}}}
\newcommand{\aeq}{\mathrel{\overset{\makebox[0pt]{\mbox{\normalfont\tiny\sffamily a}}}{=}}}
\newcommand{\beq}{\mathrel{\overset{\makebox[0pt]{\mbox{\normalfont\tiny\sffamily b}}}{=}}}
\newcommand{\capp}{\mathrel{\overset{\makebox[0pt]{\mbox{\normalfont\tiny\sffamily c}}}{\approx}}}
\def\pa{\text{pa}}
\def\G{\mathcal{G}}
\newtheorem{thm}{Theorem}

In this section, we first describe the standard constrained bidding problem, and then introduce the common RL formulation as the foundation of our paper. 

Real-time Bidding (RTB) is an important marketing channel in online advertising, which enables advertisers to gain exposure across multiple media and helps publishers to achieve monetization through the effective distribution of their traffic~\cite{defauction}. The advertisers resort to demand-side platforms (DSPs) who buy and display ads on their behalf. DSP agents repeatedly interact with billions of auctions a day and employs bidding strategies to optimize the advertisers' long-term objectives subject to various spending constraints, leading to a surge of research interest in constrained bidding~\cite{budget0,budget2,pacing1,RLB,CBRL}. 

Conventionally, the constrained bidding problem considers an RTB process comprised of auctions arriving sequentially, with the goal of scheduling bids for each auction to optimize the target utility while satisfying the relevant constraints. 
Suppose the bidding process consists of $T$ repeated auctions. At each auction triggered for an ad opportunity, the bidding agent is given (partial) information $x_i$ regarding the auction, which summarizes key details such as information about the user, the selected ad and the display context. Based on this information, the agent must decide on a bid price $b_i$. If the bid exceeds the market price $m_i=\max b_i^{-}$ (the maximal competing bids), the agent wins the auction, denoted as $\1_{b_i>m_i}$. The winning auction entails a charge of $c_i$ for ad display according to the selling mechanism prescribed by the publisher, and sends feedback about the relevant utilities, e.g., clicks, conversions. Conversely, losing auctions result only in a loose notice. In this work, we assume the online auctions adopt (or stem from) second-price auctions, holding the property of incentive compatibility~\cite{ic}. 

In the following, we focus on the ROI-Constrained Bidding (RCB) setup, which serves as a prototypical problem that can generalize to diverse advertising objectives~\cite{USCB,CBRL}. The problem of RCB aims to maximize the cumulative utility $U$ subject to a budget constraint $\1_{C\le B}$ and a return-on-investment (ROI) constraint $\1_{\roi\ge L}$:
\vspace{-.1cm}
\begin{equation}
    \max_{\mathbf{b}} U_T(\mathbf{b};\mathbf{x}), ~\text{s.t.} \quad \frac{U_T(\mathbf{b};\mathbf{x})}{C_T(\mathbf{b};\mathbf{x})}\ge L, ~B-C_T(\mathbf{b};\mathbf{x})\ge 0,\label{cb}
    \vspace{-.1cm}
\end{equation}
where the bold letters, $\mathbf{x}$ and $\mathbf{b}$, denote the sequence of auction (features) and bids, the quantity $U_T(\mathbf{b};\mathbf{x})\defeq\sum_{i=1}^T \E\lb \left. u_i \right\vert x_i\rb \1_{b_i>m_i}$ denote the cumulative utility and $C_T(\mathbf{b};\mathbf{x})\defeq\sum_{i=1}^T \E\lb \left. c_i \right\vert x_i\rb \1_{b_i>m_i}$ denote the cumulative cost.

Existing approaches on RCB have achieved promising results based on a reinforcement learning (RL) formulation, empowered by its ability in long-term planning~\cite{USCB,CBRL}. 
Following this trend, we adopt the Partially Observable Constrained MDP (POCMDP) formulation proposed in CBRL~\cite{CBRL} as the foundation of our work. In the following, we briefly summarize the main idea of the POCMDP formulation and we refer the readers to \citet{CBRL} for details.

Many leading DSPs, such as Google~\cite{Google} and Alibaba~\cite{alimama}, experience traffic throughput at the scale of billions, which poses a challenge for RL training due to the excessively lengthy decision sequences if each auction is viewed as a decision step. To mitigate this issue, CBRL adopts a distinctive perspective of the RCB problem at the aggregate level, which disentangles impression-level bid decision as a slot-wise bid ratio controlling problem with impression-level utility prediction, drawn upon the following optimal bidding theorem for constrained bidding problems~\cite{budget3,CBRL,USCB}.  
\begin{thm}
\vspace{-.05cm}
    In second-price auctions, the optimal bidding function for problem~\eqref{cb} takes the linear form of $b_i = a~u_i ~(a>0)$.\label{optb}
    \vspace{-.05cm}
    % \par\vspace{-.2cm}
    % \begin{equation}
    %     b_i = \mathbf{a}^\intercal~\mathbf{u}_i, \quad \beta>0
    % \end{equation}\label{optb}\label{thm1}
    % \par\vspace{-.4cm}
\end{thm}

This theorem states that the optimal bids (in hindsight) equals the impression value $u$ weighted linearly by a ratio $a$. Therefore, instead of treating each auction as a decision step, CBRL proposed to control the slot-wise bid ratio $a$ within a time window as a decision step, and the final bids can be computed by multiplying the bid ratio with each impression-level utility $u$.

Built upon this slot-wise ratio controlling formalism, CBRL proposed to model the bidding process as a finite-horizon episodic RL problem with $H$ time steps. Each time step $t$ represents a time window $[j_t,j_{t+1})$ containing auctions $\{x_i\}_{i\in [j_t,j_{t+1})}$. Given that the market price is only known when an auction is won, the POCMDP introduces an observation space $\O$, in addition to the full state space $\S$, in order to account for this partial observability. Both $\S$ and $\O$ contain slot-wise statistics (e.g., winning rate, ROI, total revenue and cost) but $S$ also includes information that the agent cannot observe (e.g., market price). Within this framework, the action $a\in\A$ is defined as the bid ratio scheduled for each time slot. The dynamics model $P(s^\prime,r|s, a)=\T(s^\prime\vert s, a)\cdot P(r\vert s, a)$ accounts for the transition and reward function conceptually, but the exact function mappings are unknown. 

Specifically, for the transition model, we assume partial observability of the market since the selling mechanisms and rival strategies are not transparent in sealed-bid autions. For the reward model, the step-wise reward should conceptually account for both utility and constraint violations, which involves non-trivial credit assignment to each time slot. Despite the unknown dynamics model, we can still simulate bidding environments as long as the market price is known. In this regard, past bidding logs can construct a large number of bidding environments to serve as our dataset. In addition, we can compute optimal decision sequences for each environment by solving linear programs~\cite{CBRL}, which we will denote as expert trajectories in the following passages.

In the above POCMDP formulation, we aim to find a policy $\pi$ which is a member of the policy space $\Pi:\O\times\H\mapsto \mathrm{P}(\A)$. The policy inputs past trajectories $h_t = \{o_i,a_i,r_i\}_{i=1}^{t-1}$ in addition to the current observation $o_t$, as is common practice in partially observable MDPs~\cite{xie2020deep}. The standard objective for identifying a stationary policy is to maximize the policy's value under $\M$, given by the expected cumulative reward $V(\pi;\M)=\E\lb \sum_{t=1}^H r_t \vert \pi,\M\rb$. As the bidding environment can differ from day to day, it is crucial for the bidding policy to act adaptively in varying conditions. To this end, previous methods typically adopt the following RL objective, 
% \vspace{-.1cm}
\begin{equation}
    \max_{\pi}\E_{\M} \lb V(\pi;\M)\rb,\label{erm}
    % \vspace{-.1cm}
\end{equation}
which optimizes the policy over a distribution of MDPs $p(\M)$, assuming the test distribution of environments is i.i.d. with the train distribution. Essentially, the objective embodies the principle of meta-reinforcement learning~\cite{varibad, CASTR}, aiming to meta-learn an adaptive policy that generalizes across multiple environments. \looseness-1

% \section{Methodology}
% overview: todo
\def\modelplot{
\begin{figure*}[t]
    \centering
    \includegraphics[width=\linewidth]{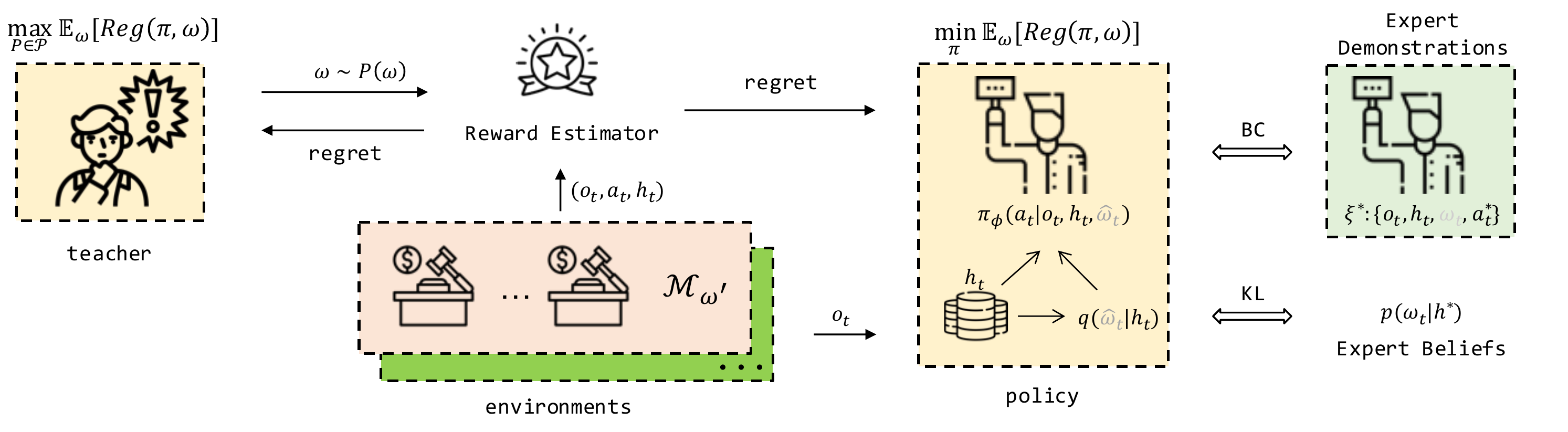}
    \vspace{-.8cm}
    \caption{Overview for MiROCL. Our method solves a differentiable game alternating between a teacher and a learner. The teacher finds a distribution $P(\boldo)\in\mathcal{P}$ of worst-case environments, and the learner meta-learns its policy $\pi$ over the given distribution of environments. In order to align with the experts in terms of causal structure, the policy $\pi$ is designed as $\pi_\phi$ conditioning on $\hat{\omega}$ obtained via the inference model $q(\hat{\omega}_t|h_t)$. Besides supervision from value maximization, the sub-policy $\pi_\phi$ and the inference model receive guidance from expert demonstrations and experts' posterior beliefs. }
    \label{fig:model}
    \vspace{-.3cm}
\end{figure*}
}
\def\scmplot{
\begin{figure}[t]
    \centering
    \includegraphics[width=\linewidth]{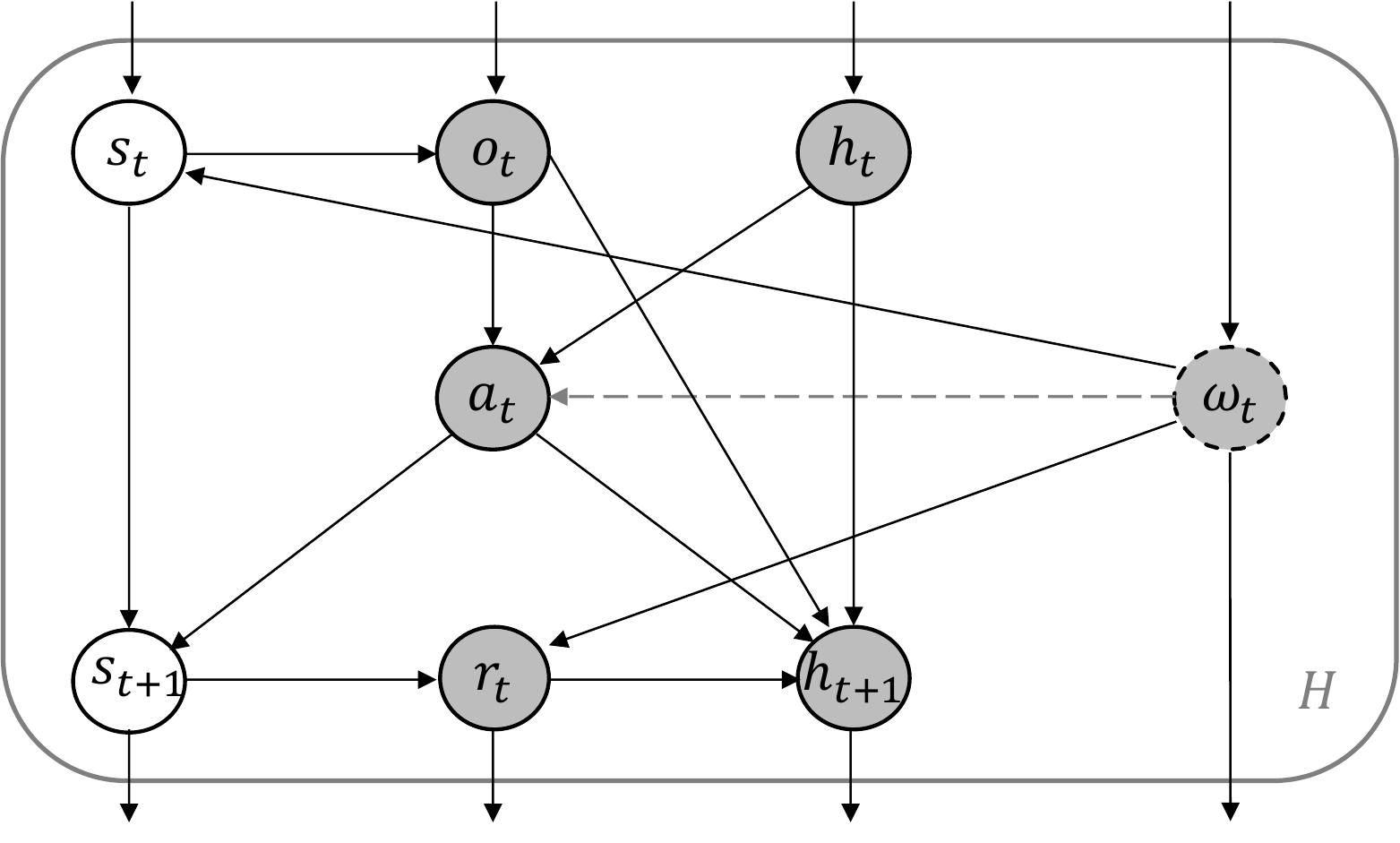}
    \vspace{-.7cm}
    \caption{Causal Diagram un-rolled at step $t$. Observed variables are shaded, while unobserved variables are not. The variable $\omega_t$ (doted edge) is observed and directed to $a_t$ in the expert's causal diagram $\G^\xi$ while unobserved with no link to $a_t$ in the policy's causal diagram $\G^\pi$. }
    \label{scm}
    \vspace{-.3cm}
\end{figure}
}
\def\ermplot{
\begin{figure}[t]
    \centering
    \includegraphics[width=\linewidth]{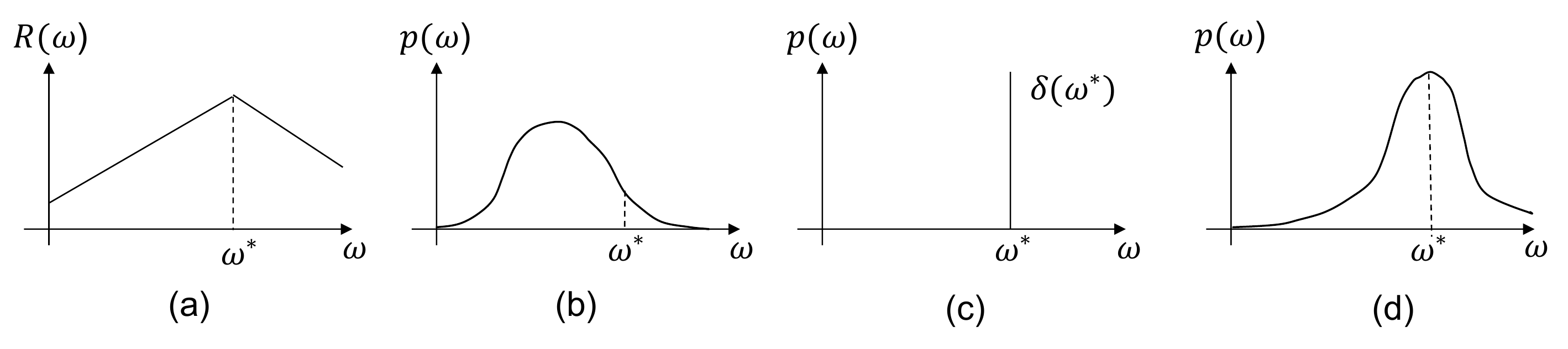}
    \vspace{-.7cm}
    \caption{ERM v.s. MiRO. (a) exemplifies $\reg(\pi,\omega)$ under different environments characterized by one-dimensional $\omega$.  (b) shows the empirical distribution of training environments. (c) shows the distribution of the test environment, which violates the iid assumption of ERM. In contrast, MiRO assumes it proportional to $\reg(\pi,\omega)$ as (d) shows. }
    \label{fig:erm_vs_miro}
    \vspace{-.3cm}
\end{figure}
}

\section{Methodology}\label{sec:miro}
% 文章注意到一个新问题：environment不仅是随机的分布，它更倾向于是对抗的。
% 环境的对抗性体现在publisher和rival bidder策略发生变化，并且变化的结果会导致我们投放效果的负向变化。两者的影响通过市场成本发生的，publisher会为了提高营收而调整其扣费，rival bidder会为了提高曝光概率提高出价从而影响竞得价格，成本的提高导致在ROI约束下，相同的花销收入会变少。
While it is well agreed recently that online ad markets dynamically changes~\cite{USCB,CBRL}, we emphasize that the bidding environment can be essentially adversarial~\cite{advbid1,advbid2} because online auctions involve multiple parties with conflicting objectives. For instance, the sellers may update their mechanism towards maximal revenue, e.g., by learning personalized reserve prices~\cite{reserve1,reserve2}, or even automatically learn mechanisms from data~\cite{regretnet,auctionlearn}. On the other hand, rival bidders can employ data-driven auto-bidding algorithms to optimize their own utilities. Moreover, the propensity of users to click an ad can vary over time due to exogenous factors, leading to incorrect utility estimations~\cite{onlearn4,repeatauctionsurvey}. Observations in real-world data also support these conjectures, as detailed in the appendix.

Unfortunately, none of the adversarial factors are directly observable to our agent, as online auctions typically seal the competing bids and the sellers have little incentive to disclose how they update their selling mechanisms. In light of this, we explore in this paper the uncharted problem of constrained bidding in adversarial environments (i.e., adversarial constrained bidding), which assumes no knowledge about how adversarial factors cause perturbations to the environment. From a game-theoretic perspective, we aim to design a bidding strategy that can effectively resist the black-box adversary in the subsequent round, leveraging the interaction history of previous rounds. 

Adversarial constrained bidding is especially challenging because the widely adopted assumption of i.i.d. train and test environments is violated, as the test environments in the adversarial setting can be purposefully manipulated towards unfavorable ones, as depicted in Fig.~\ref{fig:erm_vs_miro}. Consequently, existing works following this assumption is inapt for adversarial constrained bidding. 

To address this issue, our main insight is to \emph{align the train distribution of environments with the potential test distribution} instead of relying on the i.i.d. assumption. In what follows, we give a practical solution that realizes this insight. We first illustrate how to mathematically formulate this insight into a Minimax Regret Optimization (MiRO) framework. Then we discuss several practical issues and elaborate on how to improve upon the crude MiRO framework. \looseness-1

\subsection{The MiRO Framework}\label{sec:framework}

In this section, we first address two fundamental questions related with achieving train-test distribution alignment using collected bidding logs, and finally reach at the proposed minimax regret optimization (MiRO) framework for adversarial constrained bidding. Specifically, to achieve train-test alignment, we must first answer the following questions: what property does the adversarial setting imply about the test distribution? And secondly, how to identify the aligning train distribution given such property? 

\subsubsection{The Property of the Test Distribution.}
Through the lens of game theory, an adversary with complete knowledge of our prior strategy can perturb the environment into a worst-case scenario for this strategy. While it remains uncertain as to how such an adversary could feasibly exploit our strategy in practice, the strict condition provides us with a valuable insight into the generic adversarial setting. SSpecifically, we are led to believe that the likelihood of an environment encountered during testing is proportional to the performance a policy performs in that environment. 

To mathematically formulate this idea, we must first introduce the performance measure of policies (i.e., bidding strategies). In the adversarial setting, there are no stationary policies for all environments, so we use regret as a relative performance metric w.r.t. an oracle for each environment~\cite{prophet,onlearn1,advbid1}. Regret measures the value difference of two policies, an optimal policy (known as oracle or prophet~\cite{prophet} but we mention as the expert in the following passages), and the policy being learned:
\begin{equation}
    \reg(\pi;\M) \defeq V(\xi^\ast;\M)-V(\pi;\M) = V^\ast_\M -V(\pi;\M),
\end{equation}
where $\xi^\ast = \arg\max_{\xi\in\Xi} V(\xi;\M)$ denotes an expert policy for environment $\M$ and $V^\ast_\M$ shorthands for its cumulative value. It should be noted that, since we do not know the exact function structure of the environment, we cannot directly solve for the expert policy function that maps from any given observations to the optimal decisions. Indeed, we compute the expert trajectories based on the offline bidding logs using approximate dynamic programming, and hence the expert demonstrations conceptually requires the knowledge of anticipatory information about future dynamics. To this end, we conceptually define the expert policy space as $\xi\in\Xi:\O\times\H\times\W\mapsto \mathrm{P}(\A)$, which additionally inputs the privileged information of adversarial factors $\omega\in \W$. 

Since we must choose a representation for the test distribution, we opt for using the general energy-based distribution to express the aforementioned proportional property of the test distribution,
\begin{equation}
    P_{\text{test}}(\M) = \frac{\exp\left(\frac{1}{\alpha}\reg(\pi;\M)\right)}{Z(\pi)},
\end{equation}
where we set the regret function $\reg(\pi;\M)$ as the free energy with temperature $\alpha$, and the partition function $Z(\pi)$ serves to normalize the distribution, albeit without contributing gradients. 

\subsubsection{Identifying the Aligning Train Distribution.}
Having established the potential form of the test distribution, our focus now turns to identifying an appropriate train distribution from the collection of environments. To ensure that the aligning train distribution adheres to the available training set, we choose to project the train distribution into the set of parameterized distributions $\mathcal{P}$ (defined later in Sec.~\ref{optdiff}, which delegate the training set) in terms of the Kullback-Leibler (KL) divergence, which turns out an entropy-regularized regret maximization objective,
\begin{equation}
\begin{aligned}
    % &\reg(\pi;\M) \defeq V(\xi^\ast;\M)-V(\pi;\M).\\
    % &\\
&\min_{P\in\mathcal{P}}\D_{KL}\left(P\|P_\text{test}\right)\\
    =&\max_{P(\M)\in \mathcal{P}}\E_{\M}[\reg(\pi;\M)] + \alpha\mathcal{H}(P) +\text{const}.
    % \\
    % \text{s.t.} \quad &\pi = \arg\min_{\pi} \E_{\M}[\reg(\pi;\M)]
    % \\
    % &\min_\pi \max_{P\in\mathcal{P}} \E_{\M}[\reg(\pi;\M)] + \alpha\mathcal{H}(P) 
\end{aligned}
\end{equation}

Intuitively, the objective aims to find a distribution of environments within the training set that induces high policy regret, while adhering to the maximum entropy principle since we have no knowledge about the real adversary. The entropy regularizer controls how the distribution bias towards the worst-case environments with the temperature hyperparameter $\alpha$.  In effect, this hyperparameter anneals the interpolation between the strict adversarial setting and the iid stochastic setting, reflecting the belief of the adversarial setting. More specifically, at $\alpha=0$, the induced distribution solely focuses on worst-case scenarios, while at $\alpha\rightarrow \infty$, the induced distribution puts a uniform mass over the training set.

\subsubsection{Minimax Regret Optimization.}

Given the aligning train distribution, the policy aims to minimize its regret under this distribution. Therefore, we reach at the following (entropy-regularized) minimax regret optimization (MiRO) framework, 
\begin{equation}
    % \min_\pi \max_{\M\in \U(\M)}\reg(\pi;\M),
    \min_\pi \max_{P\in\mathcal{P}} \E_{\M}[\reg(\pi;\M)] + \alpha\mathcal{H}(P).
    \label{miro1}
\end{equation}

The MiRO framework presents a minimax game between two players, where the inner problem searches for a distribution of environments $P(\M)$ likely to align with the adversarial test conditions, and the outer problem optimizes the policy's performance under the given environments. Compared with Empirical Risk Minimization (ERM) widely adopted in previous works, which assumes small training empirical risk could generalize to small test risk, MiRO implies generalization because the policy strives to minimize the (approximately) worst-case regret that upper bounds test regret.

While MiRO appears appealing, solving such a bi-level optimization is generally intractable. Our main idea is to convert the minimax problem into a class of `differentiable games'~\cite{diffgame} so that we can resort to dual ascent~\cite{cvxopt} to search for a useful solution, as supported by generative adversarial networks~\cite{gan} and its follow-up works.\looseness=-1

\ermplot{}
\subsection{A Practical Algorithm for MiRO}\label{sec:miro_p}
In this section, our goal is to make Eq.~\eqref{miro1} a differentiable game, which is circumscribed by the unknown structure of the environment $M$ due to unobservable adversarial factors. To overcome this challenge, we propose to learn the latent representation $\omega\in\W$ of those adversarial factors from bidding logs by reconstructing the causal structure of the world model. Through world model reconstruction, we can achieve two key benefits. Firstly, since the adversarial factors  $\omega$ explain the variations of the environment, we can instead search for the distribution $p(\omega)$ supported in the learned latent space, replacing the environment $\M$ with $\omega$ in Eq.~\eqref{miro1}. And secondly, world model reconstruction establishes the map from $\omega$ to the rewards $r$, which enables the differentiation through the regret function. To this end, we can directly search for a distribution in MiRO with gradient-based optimization.

\subsubsection{Rendering a Differentiable Game.}\label{worldmodel}
To learn representations for the adversarial factors that can reflect the cause-effects on the environment, we first analyze the causal structure of the environment, and then leverage the Variational Information Bottleneck (VIB)~\cite{vib} for representation learning, which aims to learn maximally compressed representations from the input that retain maximal informativeness about the output.

\scmplot{}
We begin by describing the causal diagram~\cite{causality} (a type of augmented probabilistic graphical models~\cite{prml} used to analyze the cause-effects between random variables) of a folded time step of the bidding process. Fig.~\ref{scm} shows the cause-effects of two policies: the expert $\xi\in\Xi$ and the policy $\pi\in\Pi$ intervening with the environment. Both policies' interventions give rise to the observed variables $\{o_t,a_t,h_t\}_{t=1}^H$ and the unobserved variables $\{s_t\}_{t=1}^H$, with the causal relationships between these variables. However they differ at the variables $\{\omega_t\}_{t=1}^H$ and the cause-effects $\{a_t\rightarrow\omega_t\}_{t=1}^H$ (dotted line), since the expert conceptually knows the privileged information to give optimal decisions.

Based on these cause-effect relationships, we construct a world model including the following components: (1) the embedding model $p(\omega_t|h_t)$ that maps the history trajectory $h_t$ to step-$t$ adversary variable $\omega_t$; (2) the observation model $p(o_t,a_t,r_t|\omega_t,h_t)$ that recovers the observations $(o_t,a_t,r_t)$ from the history and adversary; (3) the latent dynamics model $p(\omega_t|\omega_{t-1}, a_{t-1})$ that models the transition in the embedding space. These probabilistic models are assumed Gaussian distributions with mean and variance implemented as neural network function approximators. For example, the embedding model takes the form of $p(\omega_t|h_t)=\mathcal{N}(\omega_t|f^\mu_\theta(h_t),f^\sigma_\theta(h_t))$.

The embedding model $p(\omega_t|h_t)$ provides the latent representation for adversarial factors, and is learned by reconstructing the observed evidence of the environment, which derives the following lower bound based on VIB: 
\begin{equation}
\begin{aligned}
\max \underset{\substack{d^\pi,d^\xi}}{\E} \lb \log p(o_t,r_t|\omega_t)\rb + \underset{d^\xi}{\E} \lb \log p(a_t|\omega_t)\rb \\-\beta \D_{KL}\left(p(\omega_t|h_t)\|p(\omega_t|\omega_{t-1},a_{t-1})\right),
\end{aligned}\label{emb}
\end{equation}
where the first two terms are the evidence of the observations, and the last term serves as a KL regularity for information compression with a hyperparameter $\beta$ controlling its strength. 

Since the adversarial factors $\omega$ explain the variations in the environments, we replace the environment $\M$ with $\omega$ in Eq.~\eqref{miro1} and instead search for a distribution $p(\omega)$ that represents the aligning train environments in the learned latent space. Meanwhile, we also achieve a tractable gradient-based search with a differentiable regret function w.r.t. $\omega$. To show this, we first write out the regret function as follows, 
\begin{equation}
    \begin{aligned}
    &\reg(\pi,\omega) = V(\xi^\ast;\MO)-V(\pi;\MO)\\
    &= {\E}
    _{d^{\xi^\ast}_\boldo}
    \lb 
    \sum_{t=1}^H \E\lb r_t\vert s_t,a_t ;\boldo\rb 
    \rb 
    - {\E}
    _{d^{\pi}_\boldo}
    \lb 
    \sum_{t=1}^H \E\lb r_t\vert s_t,a_t ;\boldo\rb 
    \rb,
    \label{regret}
\end{aligned}
\end{equation}
where $d^\pi_{\boldo}(s_t,a_t)$ (and $d^{\xi^\ast}_{\boldo}(s_t,a_t)$) denotes the policy $\pi$'s (and $\xi^\ast$'s) state-action visitations in MDP $\MO$. 

We note that, through world model reconstruction, the reward estimator $\E\lb r_t\vert s_t,a_t ;\boldo\rb$ is inherently learned as a component of the observation model $p(o_t,a_t,r_t|\omega_t,h_t)$. Specifically, we learn a neural network function approximator $r_\theta(o_t,a_t,h_t,\boldo)$ as a surrogate for the reward estimator, per the following least square objective,
\begin{equation}
    \min_\theta \E_{D}\lb \left(r^H -\sum_{t=1}^H r_\theta(o_t,a_t,h_t,\boldo)\right)^2\rb, \label{parameterization}
\end{equation}
where $D$ is the training logs containing multiple environments, and we $r^H$ denotes the episode-level reward. 
% Indeed, learning this reward estimator prepares for the gradient-based alternating optimization as the following sections show. 

\modelplot{}
\subsubsection{Optimizing the Differentiable Game.}\label{optdiff}
Finally, we reach at a differentiable game that can be optimized by an simultaneous gradient descent procedure (i.e., dual ascent) as follows, 

\begin{align}
    P^{(t)}(\boldo) &= \underset{P\in\mathcal{P}}{\arg\max}~\underset{\boldo\sim P}{\E}\lb\reg(\pi^{(t-1)},\boldo)\rb + \alpha \mathcal{H}(P),\label{miro2_env}\\
\pi^{(t)} &= \underset{\pi}{\arg\min} ~\underset{\boldo\sim P^{(t)}}{\E}\lb\reg(\pi^{(t-1)},\boldo)\rb. \label{miro2_policy}
\end{align}

Intuitively, the game alternates between the optimization of two players simultaneously -- a teacher who tutors previous policy by finding a distribution $P(\boldo)$ of worst-case environments  in the learned latent space, and a learner who meta-learns its policy $\pi$ over the given distribution of environments $P(\boldo)$.

\noindent\textbf{The worst-case tutoring step.}
To ensure that the train distribution should adhere to the empirical dataset, i.e., $P\in\mathcal{P}$, we opt for defining the set $\mathcal{P}$ based on the Wasserstein distance, which has been shown to derive convenient forms under proper assumptions~\cite{advtrn}. 

The Wasserstein distance computes the minimum cost of transform
one distribution into the other, known for the property to capture the geometry of the latent space~\cite{wgan}. Specifically, we define the Wasserstein metric $W_\kappa(\cdot,\cdot)$ with an L2-norm cost function $\kappa(\boldo,\boldo^\prime)=\|\boldo-\boldo^\prime\|_2$. Intuitively, we aim to define the set $\mathcal{P}$ as the $\rho$-neighborhood of the empirical distribution in the latent space under the Wasserstein metric. To realize this, we first denote the empirical distribution of collected environments in the latent space as $\bar{P}(\boldo)=\frac{1}{M}\sum_{i=1}^M \delta(\boldo_i)$ where $\boldo_i$ denotes the $i$-th environment. Then we define the set as $\mathcal{P} = \{P:W_\kappa\left(P,\bar{P}\right)\le\rho\}$.

Based on a dual re-formulation of Eq.~\eqref{miro2_env} (detailed in the appendix~\ref{app:teacher}), we reach at the following objective,
\begin{equation}
    \max_{P\in \mathcal{P}} \E_{\boldo} \lb \reg (\pi,\boldo)\rb = \E_{\tilde{\boldo}}\lb\max_\boldo \reg(\pi,\boldo)-\lambda\|\boldo-\tilde{\boldo}\|_{2}\rb,
\end{equation}

where $\tilde{\boldo}$ characterizes a collected environment $\M_{\tilde{\boldo}}$. In implementation, we sample a set of logged environments characterized as $\{\tilde{\boldo_i}\}_{i=1}^n$ and by gradient-based updates with stepsize $\eta$,
\begin{equation}
    \boldo^\prime \leftarrow \tilde{\boldo} +\eta \nabla_{\boldo}\lb \reg(\pi,\boldo)-\lambda\|\boldo-\tilde{\boldo}\|_{2}\rb,\label{grad}
\end{equation}
we obtain an train distribution of environments from the sampled environments, denoted as $\{\boldo^\prime_i\}_{i=1}^n$. 

\noindent\textbf{The policy improvement step.}
Given the distribution of worst-case environments $P(\boldo)=\frac{1}{n}\sum_{i=1}^n \delta(\boldo_i^\prime)$, the learner's policy improvement step according to Eq.~\eqref{miro2_policy} becomes the standard value maximization objective, as the expert value is constant w.r.t. $\pi$, 
\begin{equation}
    \min_\pi \reg (\pi;\boldo) = \min_\pi  \underset{(\tilde{\boldo},\boldo)}{\E}\lb \underset{\substack{ d_{\tilde{\boldo}}, \pi}}{\E}
        \lb 
            \sum_{t=1}^H r_\theta(o_t,a_t,h_t, \boldo)
        \rb \rb. \label{policy0}
\end{equation}

Here we note that, in implementation, each adversarial environment $\boldo$ associates with a sampled environment $\tilde{\boldo}$ due to the paired graident-based search in Eq.~\eqref{grad}. Following deep RL~\cite{sac}, we implement the policy distribution as a Gaussian distribution, with its mean and variance parameterized by neural network function approximators, i.e., $\pi(\cdot|o_t,h_t)=\mathcal{N}(\cdot|f^\mu(o_t,h_t),f^\sigma(o_t,h_t))$. 

\subsection{Causality-aware Learning with Experts}\label{sec:lowreg}
% motivation：注意到策略优化目标完全与benchmark无关。
% 作为expert demonstration，imitation RL的工作让我们有足够理由相信该策略包含一些信息能帮助策略学习。
% 没有用到benchmark
Although MiRO  is designed to minimize the policy regret (c.f. Eq.~\ref{miro1}), Eq.~\eqref{policy0} indeed degenerates to a value maximization problem without the experts' involvement. Interestingly, previous works on constrained bidding have also ignored the role of experts in their learning objectives. However, we aim to improve upon Eq.~\eqref{policy0} by also learning from expert demonstrations, as we believe the experts could entail valuable knowledge on how to optimally act in different environments. Nonetheless, It is surprising to find that the straight-forward behavioral cloning approach, which involves imitating  expert demonstrations, only results in decreased performance (Sec.~\ref{sec:bc}). To understand this issue, we view policy learning as a causal inference problem. We identify the issue of unobserved confounding that makes the policy not uniquely computable from observational data. In the following passages, we first illustrate this phenomenon and then propose a causality-based alignment strategy to remedy this issue. 

To exemplify the issue of unobserved confounding in policy learning, consider an auction environment where the selling mechanism is modified such that the winning cost is greater than the second price. In this scenario, the expert policy $\xi^\ast$ would exhibit lower bid (ratios) in comparison to second-price auctions due to the increased cost. This results in a (spurious) correlation between $a_t$ and $r_t$ (the conditioning of $(o_t,h_t)=(o_{<t},a_{<t})$ is omitted for clarity), which suggests that smaller values of $a_{t}$ associate with higher rewards $r_t$. Unfortunately, a policy learned through behavioral cloning would capture such spurious correlations as it learns only statistical dependency. As a result, such a policy would fail to generalize to environments with different selling mechanisms.

Through a causal lens, decision problems can naturally be formulated as causal queries where we aim to infer the outcome under the intervention of actions. In light of this, learning from expert demonstrations can be translated  into estimating the causal effects of interventions $do(a_t)$ on future rewards, i.e., $p(\sum_{i=t}^H r_i|o_t,h_t, do(a_t))$, using the observational data collected by the experts $\{o_i,h_i,a_i,r_i\}_{\xi}$. As shown in the causal diagram Fig.~\ref{scm}, the confounding variable $\boldo$ contributes to the causal structure of $a_t\leftarrow\omega_t\rightarrow r_t$ in the observational data, but is unobserved for the policy $\pi(a_t|o_t,h_t)$. Consequently, the conditional independence $(a_t,r_t)\!\perp\!\!\!\perp \omega_t$ is broken when $\omega_t$ unobserved, implying that the observational data presents both the causal association and the spurious correlation between $a_t$ and $r_t$. Hence, the policy cannot uniquely recover from data the desired causal query, known as the un-identifiability issue~\cite{causality}. 

To mitigate this issue, our idea is to align the causal structure of both the expert and the policy. This is achieved by conditioning the policy with an additional input $\hat{\omega}_t$, which is designed as a surrogate for the truth $\omega_t$ unavailable to the policy $\pi$ during online serving. In this regard, the policy can imitate the expert with identical causal structure so that spurious correlations is eliminated from policy learning. Therefore we adopt the following policy design, 
\begin{equation}
\vspace{-.1cm}
    \pi(a_t|o_t,h_t) = \int_{\hat{\omega}_t}\pi_\phi(\cdot|o_t,h_t,\hat{\omega}_t)\cdot
    p(\hat{\omega}_t|h_t) d\hat{\omega}_t, \label{policy}
\end{equation}
which factorizes into a sub-policy $\pi_\phi\in\Xi$ and a inference model required to infer $\hat{\omega}_t$. We note that this inference model is exactly what we have learned per Eq.~\eqref{emb}, and we aim to further leverage the guidance from expert trajectories. Therefore, we derive the following bound from minimizing policy discrepancy, 
 
\begin{equation}
\begin{aligned}
&\min_\pi \D_{KL}\left(\xi^\ast\|\pi\right) \le \min_{\pi_\phi,q} \E\lb -\log \pi_\phi(a|o_t,h_t,\hat{\omega}_t) \rb\\
    &+\beta_2\D_{KL}\left(p(\omega_t|h_t^\ast)\|p(\hat{\omega}_t|h_t)\right). \label{policyobj}
\end{aligned}
\end{equation}
The first term trains the sub-policy that inputs the inferred $\hat{\omega}_t$ to imitate the expert demonstration via behavioral cloning. The second term adds an additional KL regularity for the inference model by leveraging the posterior beliefs of the expert trajectories. The detailed derivation is included in Sec.~\ref{sec:regbound}, which suggests that the above objective indeed minimizes an upper bound on the regret.

\section{Experiments}
\def\syn{\textbf{Synthetic}}
\def\ind{\textbf{Industrial}}
\def\iid{\textbf{IID}}
\def\ood{\textbf{OOD}}
\def\pid{\textbf{PID}}
\def\cbrl{\textbf{CBRL}}
\def\uscb{\textbf{USCB}}
\def\indplots{
\begin{figure*}[t]
    % \vspace{-.2cm}
    \centering
    \begin{subfigure}[b]{0.3\textwidth}
         \centering
         \includegraphics[width=\textwidth]{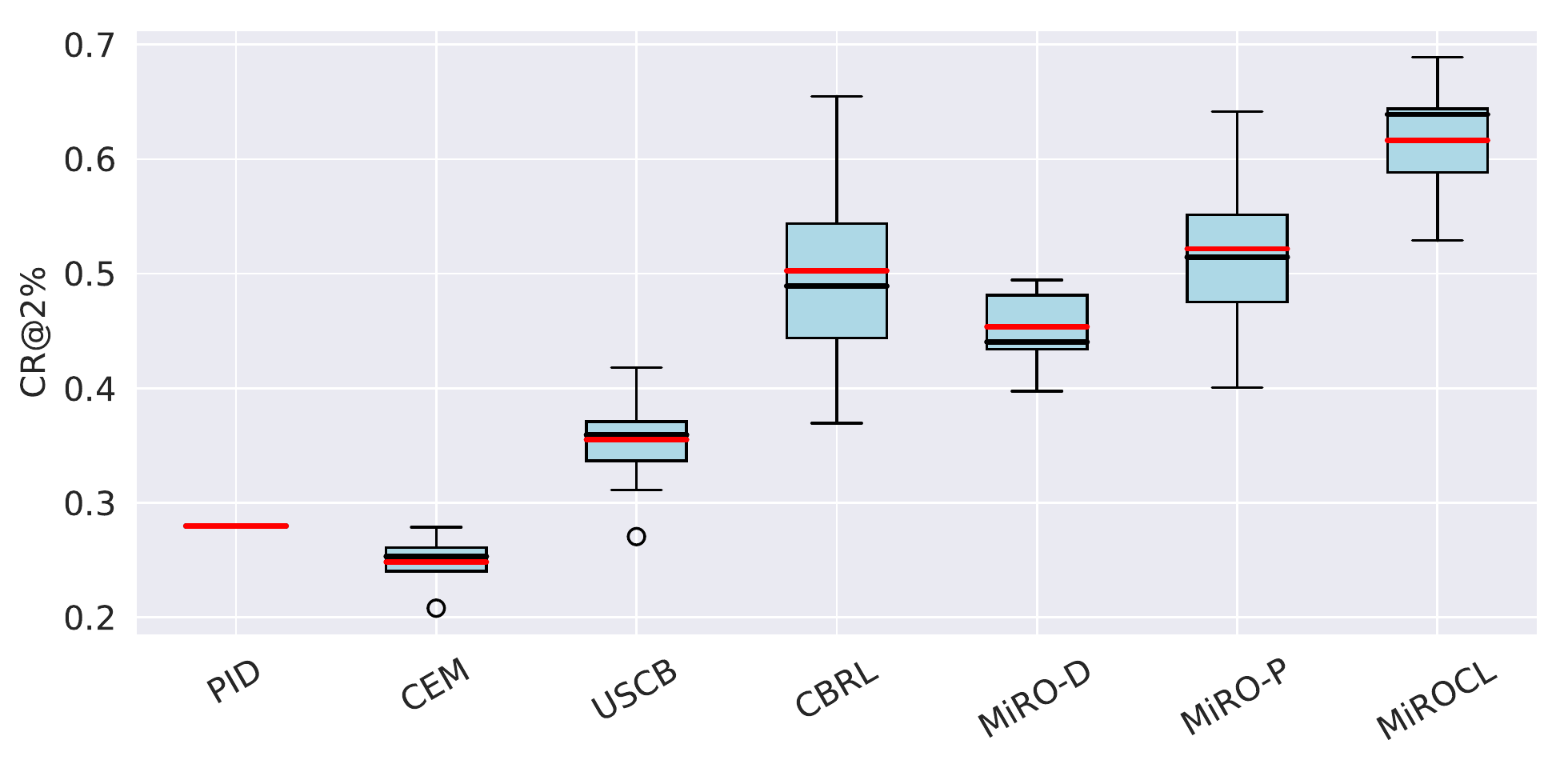}
        %  \caption{$y=x$}
        % \caption{NAPR}
        %  \label{fig:y equals x}
     \end{subfigure}
     \hfill
     \begin{subfigure}[b]{0.3\textwidth}
        \centering
        \includegraphics[width=\textwidth]{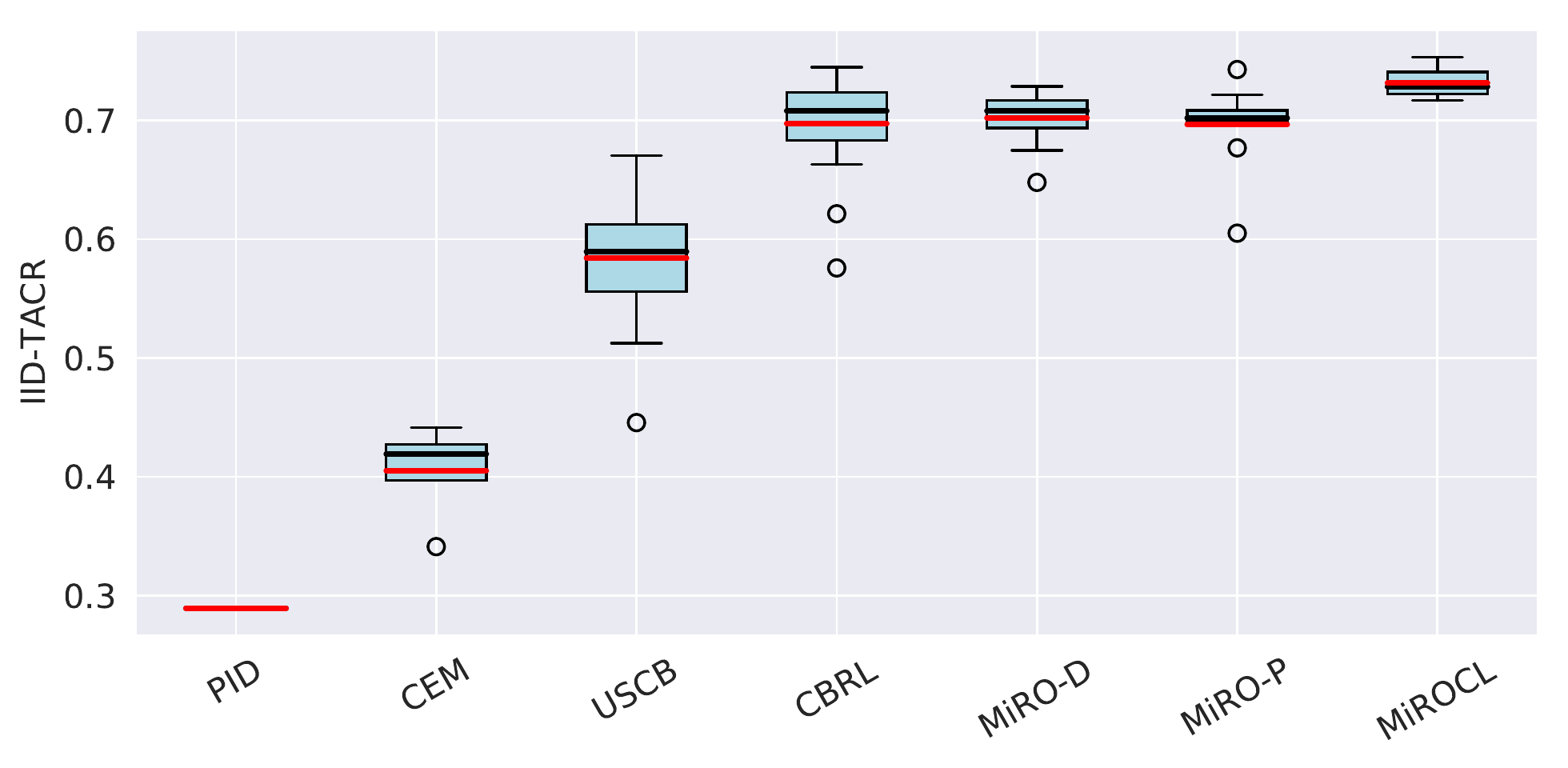}
       %  \caption{$y=x$}
       % \caption{CSR}
       %  \label{fig:y equals x}
    \end{subfigure}
    \hfill 
    \begin{subfigure}[b]{0.3\textwidth}
        \centering
        \includegraphics[width=\textwidth]{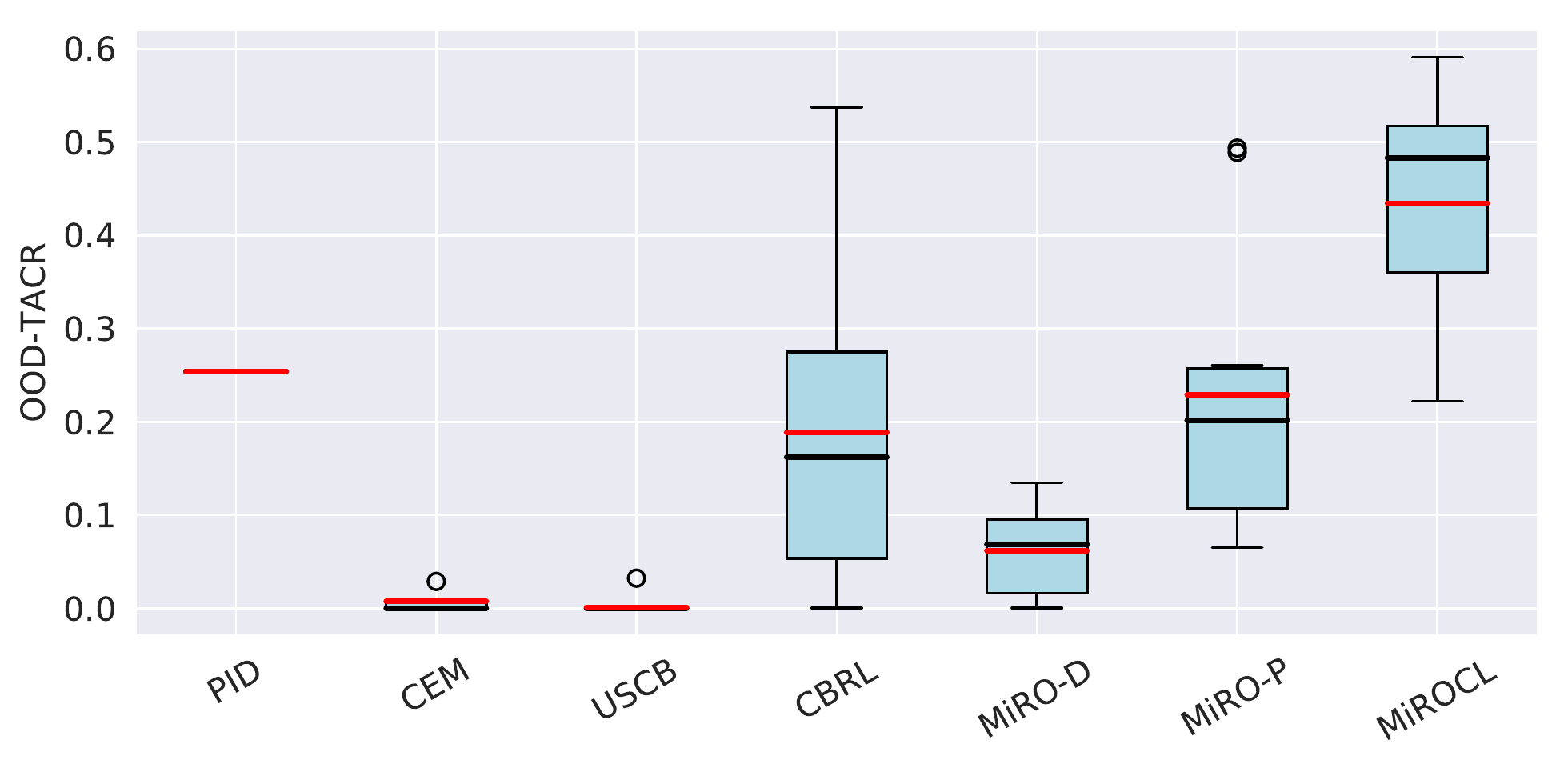}
       %  \caption{$y=x$}
       % \caption{ANDR}
       %  \label{fig:y equals x}
    \end{subfigure}

    \vspace{-.2cm}
    \caption{\small The results of CR@2\% (Left), IID-TACR (Middle), and OOD-TACR (Right) in the \ind{} dataset are shown above. Each boxplot shows the average (red) and median (black) results of 20 independent repeated runs.}
    \label{fig:ind_box}
    \vspace{-.2cm}
\end{figure*}

}
\def\indtable{
\begin{table}[t]
    \centering
    \caption{Median test scores on \ind{} which features real-world adversarial conditions. \small The reported mTACR and mCR scores take a median across $20$ random trials for all models. TACR is computed with max tolerance level $2\%$ and payoff rate $5\%$, and $CR@2\%$ shows the un-discounted CR at tolerance level $2\%$. IID-mTACR and OOD-mTACR reports detailed results on two sets, the in-distribution set and the out-of-distribution set.}
    \vspace{-.2cm}
    \begin{tabularx}{\linewidth}{cccccc}
    \toprule
        ~ & mTACR & mCR@2\% & IID-mTACR & OOD-mTACR \\ %\hline
    \midrule
        MiROCL & 0.6359 & 0.6391 & 0.7285 & 0.4833 \\ %\hline
        MiRO-P & 0.5015 & 0.5146 & 0.702 & 0.2013 \\ %\hline
        MiRO-D & 0.4337 & 0.4406 & 0.7079 & 0.0684 \\ %\hline
        CBRL & 0.4793 & 0.489 & 0.708 & 0.1622 \\ %\hline
        USCB & 0.3537 & 0.359 & 0.5895 & 0.0 \\ %\hline
        CEM & 0.2515 & 0.2529 & 0.4192 & 0.0 \\ %\hline
        PID & 0.2753 & 0.2795 & 0.2894 & 0.2542 \\ %\hline
    \bottomrule
    \end{tabularx}
    
    \label{exp:ind}
    \vspace{-.4cm}
\end{table}

}
\def\syntable{
\begin{table}[t]
    \centering
    \caption{Median scores on \syn{}. \small The reported mTACR and mCR scores take a median across $20$ random trials for all models. We also report results on two types of auction formats.}\label{exp:syn7}
    \vspace{-.2cm}
    \begin{tabularx}{\linewidth}
    {cccccc}
    % {>{\raggedright\arraybackslash}X  >{\centering\arraybackslash}X  >{\centering\arraybackslash}X>{\centering\arraybackslash}X >{\centering\arraybackslash}X >{\centering\arraybackslash}X }
    \toprule
        ~ & mTACR & mCR@2\% & GSP-mTACR & MIX-mTACR \\ 
        \midrule
        MiROCL & 0.8094 & 0.8114 & 0.7932 & 0.8098 \\ %\hline
        MiRO-P & 0.7231 & 0.7469 & 0.7307 & 0.7116 \\ 
        % \hline
        CBRL & 0.6856 & 0.7003 & 0.696 & 0.6793 \\ %\hline
        USCB & 0.5171 & 0.5304 & 0.5114 & 0.5256 \\ %\hline
        CEM & 0.5811 & 0.6094 & 0.5972 & 0.5438 \\ %\hline
        PID & 0.3343 & 0.3556 & 0.3728 & 0.2766 \\ %\hline
        \bottomrule
    \end{tabularx}
    \vspace{-.4cm}
\end{table}

}

% miro: worst-case optimal, better generalization (ood->iid)
% low-regret learning: higher test competitive ratios 
In this work, we propose a Minimax Regret Optimization (MiRO) framework for adversarial constrained bidding, with a practical algorithm for end-to-end optimization. In addition, we are the first to advocate policy learning with expert demonstrations, which enhances MiRO into \emph{MiROCL} (MiRO with Causality-aware reinforcement Learning). Therefore, we aim to examine the following questions in the experiments. 
\begin{itemize}[leftmargin=*]
    \item \textbf{Q1 (Comparison with Prior Works)}: How does the proposed method empirically perform versus prior methods in black-box adversarial environments? 
    \item \textbf{Q2 (Ablation)}: The effectiveness of each component proposed? 
    % \item \textbf{Q3}: How different methods behave in an adversarial environments with variations in the selling mechanisms?
\end{itemize} 
For these questions, we use an industrial dataset, denoted as \ind{}, which presents real-world adversarial situations.  Since adversarial factors are entangled in real-world data, we also create a synthetic dataset, denoted as \syn{}, which involves varying selling mechanisms of known structure. We put details about the dataset and the implementation in the appendix, and the dataset and code is publicly available at \href{https://github.com/HaozheJasper/MiROCL}{\textcolor{blue}{https://github.com/HaozheJasper/MiROCL}}.

% \noindent\textbf{Dataset}
\subsection{Experimental Setup}
\subsubsection{Dataset.} 
We use two datasets in the experiments. The \ind{} dataset is collected from the Alibaba display advertising platform, including $80$ days of bidding logs with each day $2$ million requests on average. Each request $x_i$ includes: the market price $m_i$, the utility estimations $\E\lb u_i|x_i\rb$ and the real stochastic feedback $u_i$. Contextual features are not used in this work as we assume estimations $\E\lb u_i|x_i\rb$ are pre-computed. As market price $m_i$ is not revealed in lost auctions in the RTB system, we use a special strategy that bids a price as high as possible to obtain the market price. Hence, \emph{we consider each bidding log in the industrial dataset represents a distinct bidding environment}. For Q1, we split the industrial dataset into the first $60$ days and the last $20$ days, based on our observation that the two sets differ in their market dynamics and/or value distributions (Fig.~\ref{fig:dataplot}). Thirty days are sampled from the in-distribution set to form the training set, the remaining $30$ days from the set form the \textbf{IID} test set and the last $20$ days as the \textbf{OOD} test set. 

The \syn{} dataset is synthesized based on the public synthetic dataset AuctionGym~\cite{jeunen2022learning}. In contrast to the relatively small-scale AuctionGym, our synthetic dataset includes $80$ days of bidding logs with each day $10$ million impressions, designed for research on constrained bidding with expert strategies. To simulate constrained bidding in adversarial environments similar to the industrial data, we assume that the real-world black-box auctions can be approximated by a format of linearly mixed second-first price auctions, i.e., the cost $c_i = k\cdot b_i + (1-k)\cdot m_i$ is a linear combination of bid $b_i$ and market price $m_i$, with possibly dynamic ratio $k$. The assumption owes to the data insights that the charge $c_i$ on some media channels is dependent on the bid $b_i$, leading to the observation that the winning probability of the same traffic distribution varies as the bids change (the charge of the same traffic distribution does not change as the bids vary in second-price auctions, and so does the winning probability). Consequently, for synthetic experiments we simulate a dynamic mixed second-first price auction environment to examine the effect of algorithms in the adversarial setting. In this case, the train set includes $10$ days of GSP bidding logs, and $20$ days of the dynamic mixed auction with randomly sampled ratio $k\in(0,1)$, whereas the test set includes $20$ days of GSP and $30$ days of randomly sampled mixed auction logs. 
% We put further descriptions about the datasets in the appendix, and we will release datasets to promote research on the adversarial constrained bidding problem.

% \noindent\textbf{Evaluation Protocols}
\subsubsection{Evaluation Protocols}
We use the \emph{competitive ratio} (CR) to evaluate methods in our experiments. CR is the ratio of the policy value versus the expert value, which directly reflects the online regret. Moreover, we introduce a notion of \emph{tolerance} in our evaluation, and it is motivated by real practices in which we can tolerate the policy violating the constraints if it gains a sizeable payoff. Specifically, we define the metric of the average tolerance-aware comparative ratio (TACR) on a dataset of $N$ days, with a pre-defined max tolerance $\gamma$ and baseline payoff rate $\zeta$ as follows,
\begin{equation}
    \text{TACR} = \frac{1}{N}\sum_{i=1}^N \frac{U(i)/(1+\zeta)^{\lambda(i)}\cdot \1_{\roi(i)\ge L(1-\gamma) }}{U^\ast(i)},
\end{equation}
where we use abbreviations $U(i) = U_T(\pi,\mathbf{x}^i)$ to denote the cumulative utility obtained by $\pi$ for request sequence $\mathbf{x}^i$ of day $i$. Similarly, $C(i)$ denotes the total cost, and $U^\ast(i)$ denotes the benchmark value. 

We compute $\lambda(i)=\max\{\text{ceil}(\max\{1-\roi(i)/L,0\}), \gamma\}$ as the tolerance level of $\pi$'s solution for day $i$. Intuitively, this quantity measures the competitive ratio allowing the violation of constraints within a maximum tolerance of $\gamma$, and the values will be discounted by a baseline payoff rate $\zeta$ if it violates the constraints.  In particular, we set $\gamma=2\%, \zeta=5\%$ in our experiment, which indicates that we consider it worthy to exchange each $1\%$ drops in ROI for $5\%$ increase in utility, within the max tolerance of violation $2\%$. In addition, we also show the competitive ratio (CR) at the max tolerance level $2\%$, which considers solutions that violate constraints below $2\%$ as feasible. We denote the metric as $CR@2\%$. 

\indplots{}
\subsection{Empirical Results}\label{sec:results}

% \noindent\textbf{Prior Methods}
\subsubsection{Comparison with Prior Methods.}
This work aims to present the challenge of constrained bidding in adversarial environments, for comparisons, we select representative methods that can or can be adapted to handle the constrained bidding problem~\eqref{cb} with both budget constraint and ROI constraint: (1) The PID control method~\cite{PID} and the cross-entropy method CEM~\cite{CEM} are viewed as online learning approaches; (2) USCB~(\citeyear{USCB}) and CBRL~(\citeyear{CBRL}) are two recently proposed RL-based methods trained using collected bidding logs. Among them, CBRL establishes with the ERM principle and learns without expert demonstrations. 

% \noindent\textbf{Comparison with prior methods.}
% \subsubsection{Comparison with prior methods.}
The evaluation results on the industrial dataset and the synthetic dataset are shown in Tab.~\ref{exp:ind} and Tab.~\ref{exp:syn7}. We empirically show that the proposed MiROCL performs the best on both datasets in terms of all performance metrics. For clarification, we note while the TACR result for one independent run shall be between its IID-TACR and OOD-TACR, mTACR are not necessarily between IID-mTACR and OOD-mTACR scores because they report the median over $20$ runs. Besides, mTACR is close to mCR@2\% for some models (e.g., MiROCL), which implies that the models likely enjoy high payoff when violating constraints (so they are less discounted and is close to mCR@2\%). In addition, some models (that follows the batch learning paradigm) show similar performance in the GSP and synthesized adversarial selling mechanisms in Tab.~\ref{exp:syn7}, because the models are trained on the joint set of different aution formats, leading to an averaging effect on different auction formats. 
The results of competing methods are as follows:

\begin{itemize}[leftmargin=*]
    \item \pid{}~\cite{PID} operates based on real-time feedback control. Tab.~\ref{exp:ind} shows that PID preserves relatively stable performance in OOD environments. However, we observe that PID performs worse in dynamic mixed second-first price auctions than in GSP (with GSP-mTACR 0.3817 versus MIX-mTACR 0.2837 in Tab.~\ref{exp:syn7}), possibly because PID is not so anticipatory as to shade some of its bid which is transformed into its cost charged by the publisher. 
    \item \textbf{CEM}~\cite{CEM} is a zeroth order stochastic optimization method supposedly not affected by the distributional shift. However, we observe that the hyper-parameters of CEM are sensitive to drastic distributional shifts (OOD scores $0$ in Tab.~\ref{exp:ind}). For example, the optimal bid ratios in OOD environments can lie beyond the touch of CEM's elite distribution (Fig.~\ref{fig:dataplot}). The algorithm achieves more stable performance in the synthetic dataset (Tab.~\ref{exp:syn7}), because the train and test conditions are i.i.d. 
    \item \uscb{}~\cite{USCB} adopts a Monte-Carlo value estimation-based actor-critic RL method with a soft reward function.  Tab.~\ref{exp:ind} shows that USCB fails to generalize in OOD environments, highly likely due to the unobserved confounding issue. In i.i.d. test conditions (Tab.~\ref{exp:syn7}), USCB performs even worse than CEM. This is because USCB's overall performance is largely determined by the hyperparameter controlling the utility-constraint trade-off. Since different selling mechanisms require a different configuration of such trade-off, a single static hyperparameter will lead to sub-optimal performance over the whole dataset. 
    \item \cbrl{}~\cite{CBRL} proposed the  POCMDP formulation that lays the foundation of our work. There are two major differences between CBRL and MiROCL. Firstly, CBRL adopts the ERM principle following the objective of Eq.~\eqref{erm}, which preserves no generalization guarantee in the adversarial setting. Secondly, CBRL do not consider including expert demonstrations for policy learning.  CBRL performs better than USCB in the challenging OOD environments in terms of OOD-mTACR, because CBRL employed a Bayesian mechanism that aims to adapt to the environment. However, CBRL still performs worse than our method, since MiRO potentially aligns the train and test distribution and can thus generalize better. In i.i.d. train and test conditions (Tab.~\ref{exp:syn7}), CBRL outperforms USCB because it implicitly learns to infer the utility-constraint trade-off for different environments. However, The proposed method still outperforms CBRL mainly because MiROCL has distilled knowledge from the experts.
\end{itemize}

% \sotaplot{}

\indtable{}

% \noindent\textbf{Q1: Does MiRO promote generalization?}
\subsubsection{Ablation Study.}\label{sec:bc}

We first recap the design and components proposed in our methodology. 
\begin{itemize}[leftmargin=*]
    \item ERM v.s. MiRO. In Sec.~\ref{sec:framework}, we pointed out the major pitfall of ERM that the i.i.d. assumption is broken in the adversarial setting. To remedy this, we propose a practical MiRO algorithm in Sec.~\ref{sec:miro_p} to achieve an elaborated train-test alignment. We choose the state-of-the-art method, \textbf{CBRL}, which follows the ERM principle as a representative of ERM-based methods, and we compare it with the practical MiRO algorithm, denoted as \textbf{MiRO-P}.
    \item MiRO v.s. MiRO with behavioral cloning. In Sec.~\ref{sec:lowreg}, we aim to improve upon MiRO by including explicit learning from expert demonstrations due to the belief that experts entail valuable knowledge about making optimal decisions in different environments. We first examine the straight-forward idea of behavioral cloning, which requires the policy to imitate the expert demonstrations following the Maximum Likelihood Estimation (MLE) principle. This approach, denoted as \textbf{MiRO-D}, shows surprising performance degradation, which motivates the proposed causality-aware alignment strategy. 
    \item MiRO v.s. MiROCL. Based on a causal analysis, we identify the issue of unobserved confounding that fails the straight-forward MLE principle. To address this issue, we proposed a causality-aware approach that disentangles the policy to a sub-policy that mimics the expert's causal structure and the inference model. This overall method is denoted as \textbf{MiROCL}.
\end{itemize}

We then give a summarization of the empirical results.
\begin{itemize}[leftmargin=*]
\item ERM v.s. MiRO. As shown in Tab.~\ref{exp:ind}, MiRO-P has better overall performance than CBRL. While CBRL and MiRO-P has comparable IID performance, MiRO-P shows better average performance and stability than CBRL in terms of real-world OOD situations, according to the boxplots in Fig.~\ref{fig:ind_box}. This is because the MiRO framework allows the policy to train under a distribution more likely to align with the test distribution, thus resulting in better average performance. However, we also witnessed that MiRO is still limited in stability (large variance in boxplots) due to the unpreditability of test conditions and stochasticity during training, implying a interesting direction for future research. 
\item MiRO v.s. MiRO with behavioral cloning. Similar to the comparision with ERM, MiRO-P has better overall performance than MiRO-D mainly conducive to better OOD performance. In real-world OOD situations, MiRO-D shows worse average performance with higher stability as compared with MiRO-P according to Fig.~\ref{fig:ind_box}. This is because behavioral cloning involves with the unobserved confounding issue, which learns the spurious statistical dependency that might not generalize under distributional shifts. We conjecture that MiRO-D shows better stability due to the memorization effect of behavioral cloning, implying that the decisions are less adaptive to different environments. 
\item MiRO v.s. MiROCL. As shown in Tab.~\ref{exp:ind}, MiROCL has significant improvement over MiRO-P in terms of both IID and OOD performance, with  $21\%$ increase in OOD-mTACR in particular. The IID performance improves mainly due to the effective distillation from expert demonstrations. The OOD performance improves because MiROCL follows the causal structure of the expert policy and leverages the inference model that adaptively infer the privileged information in different environments. Notably, Tab.~\ref{exp:syn7} also shows that MiROCL improves MiRO-P by a large margin  ($12\%$) in adversarial selling mechanisms, indicating that expert demonstrations effectively guide the policy toward optimality. 

\end{itemize}

\syntable{}

\section{Related Work}
\textbf{Constrained Bidding.} We discuss related works on the second-price auctions. A majority of research on constrained bidding follows the iid assumption or the ERM principle. Among them, most works focus on bidding with at most a budget constraint (c.f. \cite{optsurvey} for a survey), while some works~\cite{PID,USCB,CBRL} further propose to deal with the more challenging cost-related constraints, i.e., ROI-like constraints. Our work investigate the problem of ROI-Constrained Bidding (RCB), based on the POCMDP formulation in CBRL~\cite{CBRL}. 

\noindent\textbf{Connections with adversarial learning in repeated auctions.} Recently, a few recent works~\cite{advbid1, advbid2} discuss the adversarial learning to bid problem. \cite{advbid1} discusses an online learning approach in adversarial first-price auctions and deals with no constraints, which is orthogonal to our work. \cite{advbid2} investigate the scenario with adversarial sellers, and assumes the seller adopts data-drive mechanism~\cite{regretnet}. In reality, however, there are multiple sources of adversarial factors none of which are observable to the agents. To address this limitation, we explore the prior-free adversarial setting. 

\noindent\textbf{Connections with the minimax game formulation.} 
The formulation of a minimax game is also seen in generative adversarial networks~\cite{gan, wgan}, regret analysis in online learning~\cite{advbid1, onlearn1,onlearn3}, distributionally robust optimization (DRO) in supervised learning ~\cite{dro1,dro2} and adversarial training~\cite{advtrn}.  GANs aims to achieve realistic generation that links minimizing distribution discrepancy with the minimax objective. DRO and adversarial training relates with robust generalization under distributional shift and adversarial attacks. Regret analysis aims to provably bound the performance gap under the worst online conditions. In particular, previous online learning approaches for bidding~\cite{onlearn1,onlearn2,onlearn3,onlearn4} are typically limited to small-scale problems and require knowledge about the market.  Our approach can be seen as combining the merits of minimax-optimality principle of  online learning and robustness considerations of offline learning, but is motivated by the insight of train-test distribution alignment. 

\vspace{-.2cm}

\section{Conclusion}
In this work, we explore an uncharted problem of constrained bidding in adversarial environments without knowledge about how adversarial factors perturb the environments.  Previous methods on constrained bidding typically rely on the Emipirical Risk Minimization (ERM) principle which is violated in the adversarial setting. To address this limitation, we propose a Minimax Regret Optimization (MiRO) framework, which interleaves between a teacher identifying the aligning train distribution and a learner optimizing the policy under the given distribution of environments. To make the minimax problem tractable, we renders a differentiable game by variational learning the representation of adversarial factors by reconstructing the causal structure of the world model, and optimizes the differentiable game via dual gradient descent. In addition, we are the first to incorporate expert demonstrations for policy learning. We identify the unobserved confounding issue that fails the straight-forward idea of behavioral cloning from experts, and develop a causality-aware approach that aims to mimic the causal structure of the expert policy and distill knowledge from expert demonstrations. Empirical results on both large-scale industrial and synthetic dataset show that our method, MiROCL, outperforms prior methods more than $30\%$.

\bibliographystyle{ACM-Reference-Format}
\balance
\bibliography{paper-ref}

\clearpage
\appendix
\def\sotaplot{
\begin{figure}[t]
    \centering
    \includegraphics[width=\linewidth]{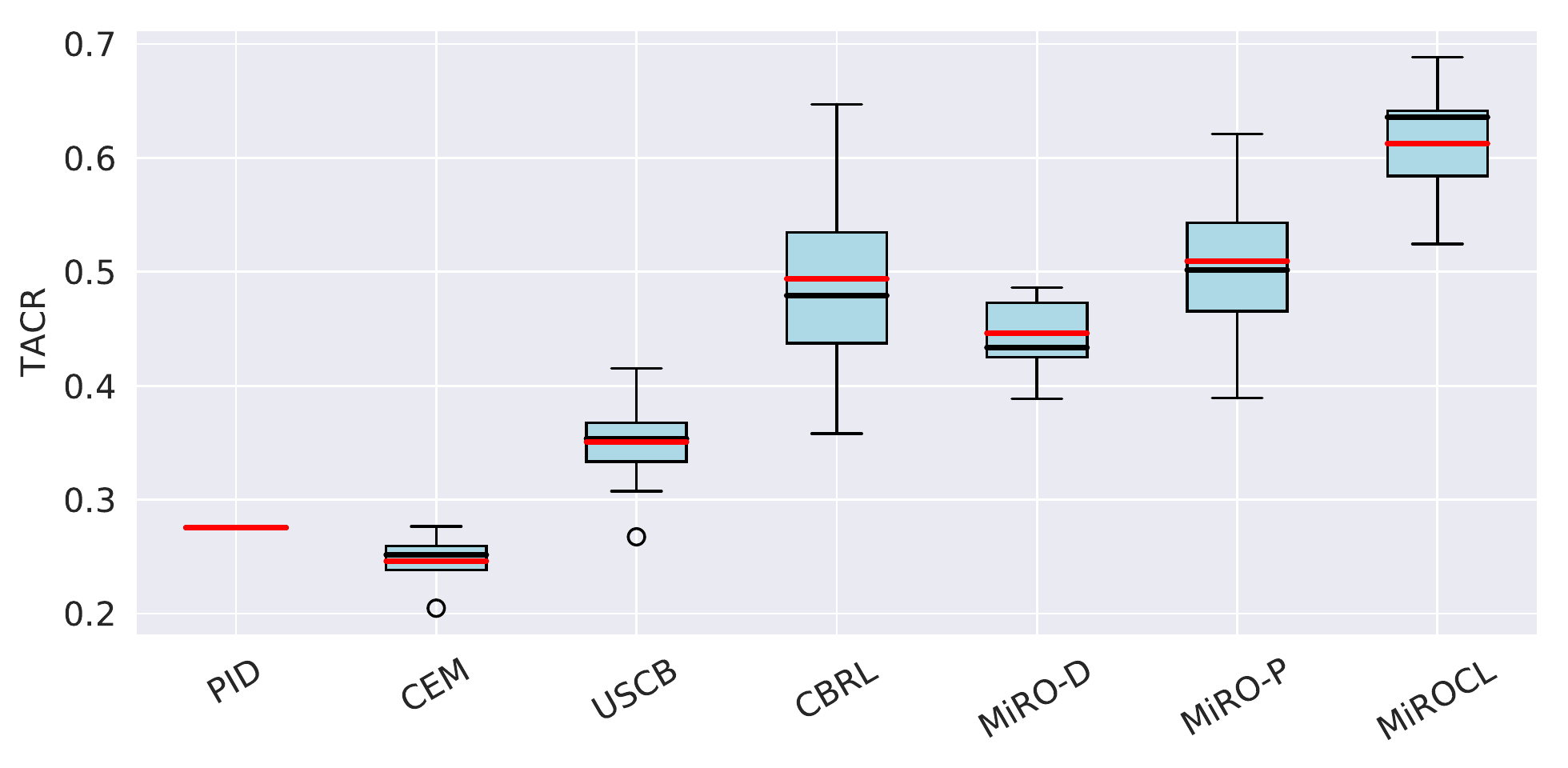}
    \caption{TACR result on the \ind{} dataset. Each boxplot shows the median (red bar) and mean (black bar) scores over $20$ random trials. }
    \vspace{-.4cm}
    \label{fig:sota_yewu}
\end{figure}}
\def\dataplot{
\begin{figure}[!t]
        \centering
        
        \begin{subfigure}[c]{0.46\linewidth}
             \centering
             \includegraphics[width=\textwidth]{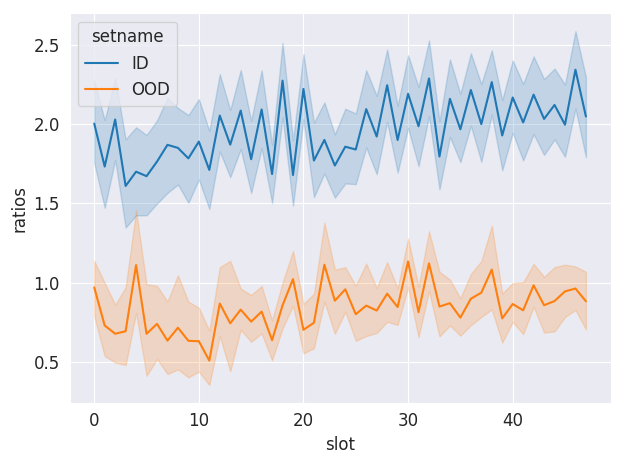}
            %  \caption{NAPR}
            %  \label{fig:y equals x}
         \end{subfigure}
         \hfill
         \begin{subfigure}[c]{0.46\linewidth}
             \centering
             \includegraphics[width=\textwidth]{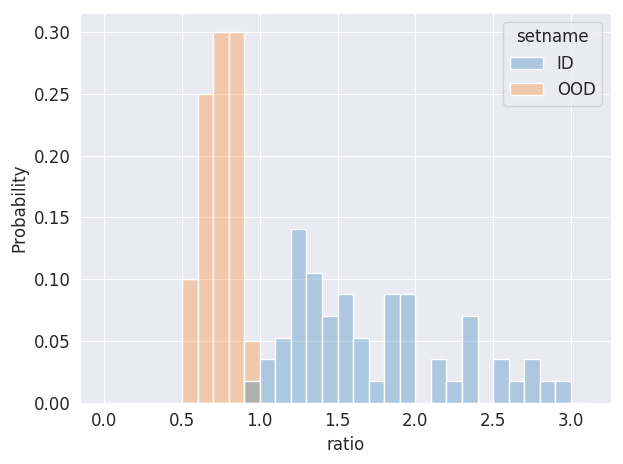}
            %  \caption{CSR}
            %  \label{fig:y equals x}
         \end{subfigure}
        %  \hfill
        %  \begin{subfigure}[b]{0.3\textwidth}
        %      \centering
        %      \includegraphics[width=\textwidth]{sections/figs/yewu_ANDR.png}
        %      \caption{the conditional revenue improvement over oracle ANDR}
        %     %  \label{fig:y equals x}
        %  \end{subfigure}
        \vspace{-.3cm}
        \caption{The distribution shift of the \ind{} dataset. (Left) the average slot-wise expert policy on the IID and OOD set. (Right) The distribution of the bid ratio of a day-wise expert policy.\looseness=-1 }
        \label{fig:dataplot}
        \vspace{-.3cm}
    \end{figure}
}
\def\synplot{
\begin{figure}[t]
    \centering
    \includegraphics[width=\linewidth]{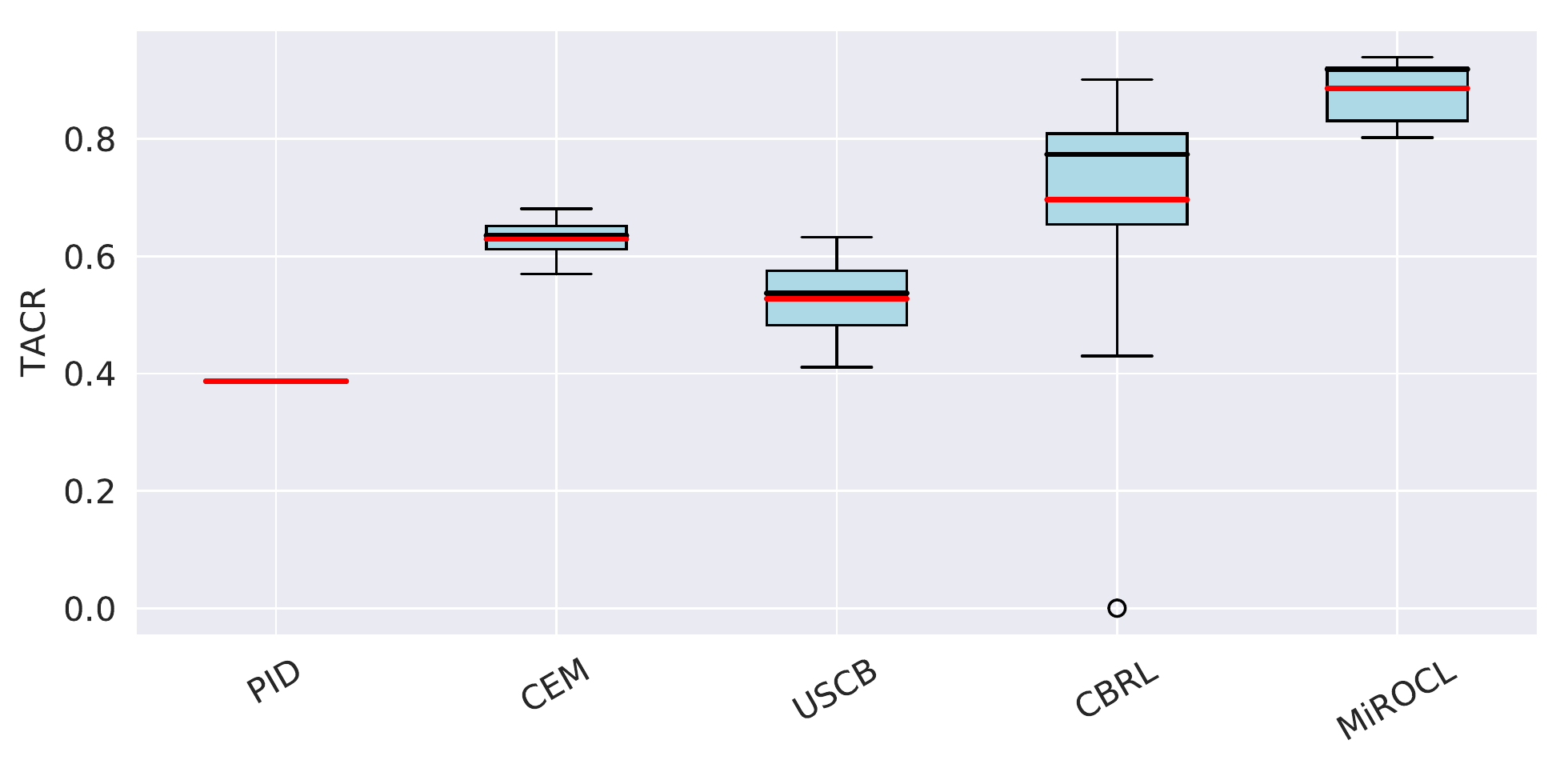}
    \vspace{-.6cm}
    \caption{TACR result on the \syn{} dataset. Each boxplot shows the median (red bar) and mean (black bar) score over $20$ random trials. }
    \vspace{-.2cm}
    \label{fig:syn_tacr}
\end{figure}}
\def\plots{
\begin{figure*}[t]
    \vspace{-.3cm}
    \centering
    % \begin{subfigure}[b]{0.3\textwidth}
    %      \centering
    %      \includegraphics[width=\textwidth]{sections/figures/all_models_yewu_CR@2.pdf}
    %     %  \caption{$y=x$}
    %     % \caption{NAPR}
    %     %  \label{fig:y equals x}
    %  \end{subfigure}
    %  \hfill
    %  \begin{subfigure}[b]{0.3\textwidth}
    %     \centering
    %     \includegraphics[width=\textwidth]{sections/figures/all_models_yewu_IID-TACR.pdf}
    %    %  \caption{$y=x$}
    %    % \caption{CSR}
    %    %  \label{fig:y equals x}
    % \end{subfigure}
    % \hfill 
    % \begin{subfigure}[b]{0.3\textwidth}
    %     \centering
    %     \includegraphics[width=\textwidth]{sections/figures/all_models_yewu_OOD-TACR.pdf}
    %    %  \caption{$y=x$}
    %    % \caption{ANDR}
    %    %  \label{fig:y equals x}
    % \end{subfigure}

    \begin{subfigure}[b]{0.3\textwidth}
        \centering
        \includegraphics[width=\textwidth]{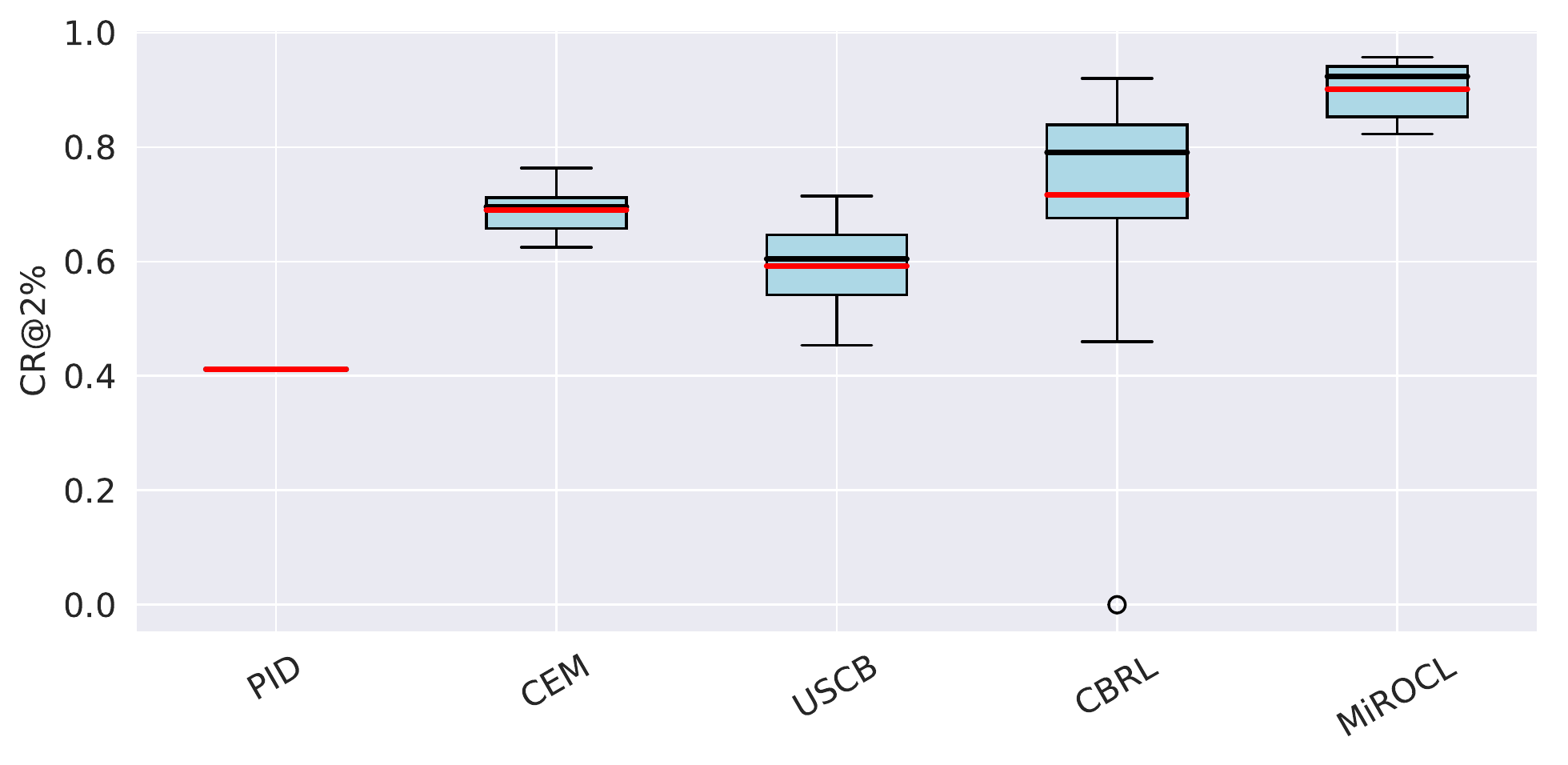}
       %  \caption{the overall performance NAPR}
       %  \label{fig:y equals x}
    \end{subfigure}
    \hfill
    \begin{subfigure}[b]{0.3\textwidth}
        \centering
        \includegraphics[width=\textwidth]{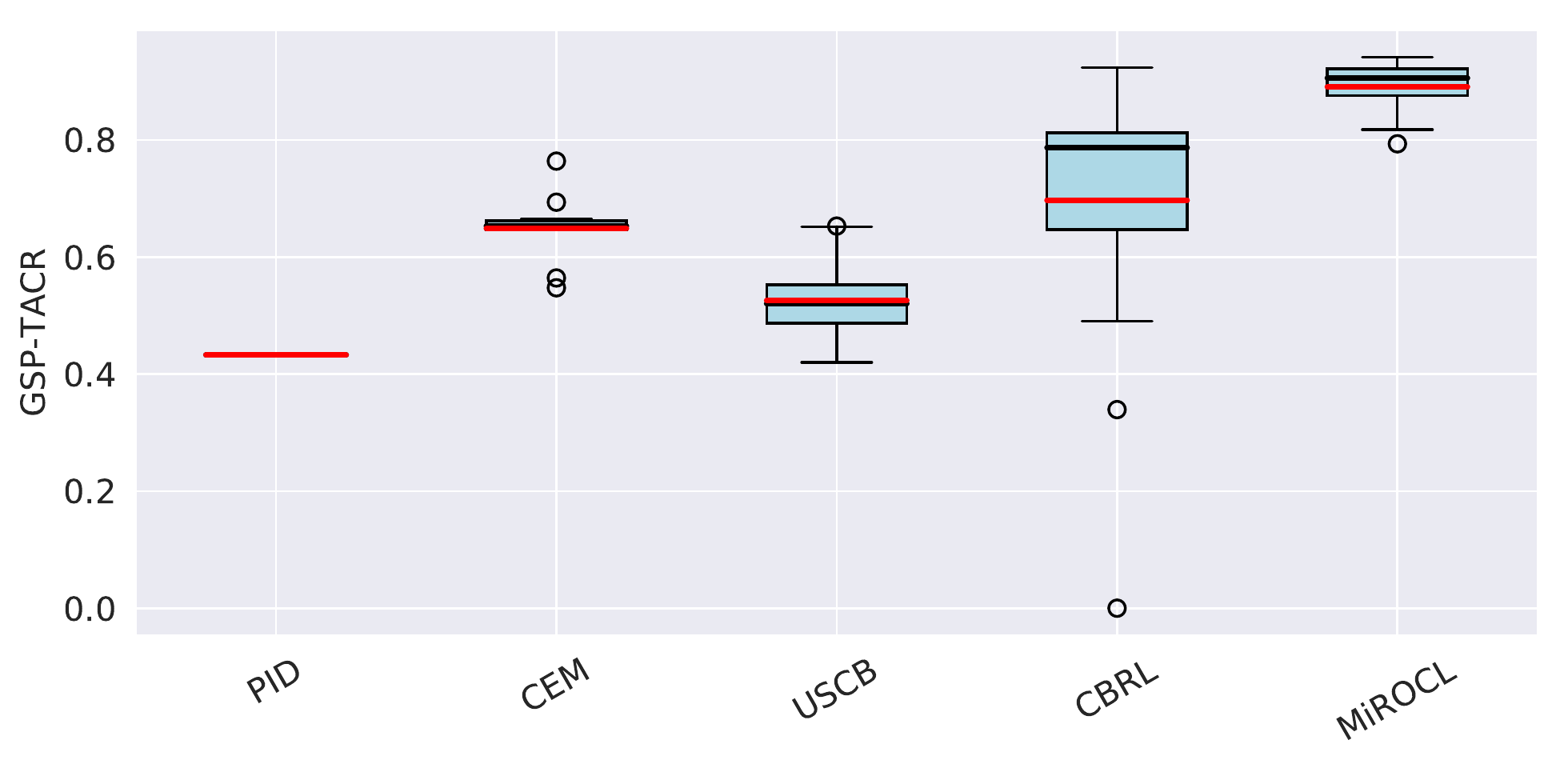}
       %  \caption{the constraint satisfaction ratio CSR}
       %  \label{fig:y equals x}
    \end{subfigure}
    \hfill
    \begin{subfigure}[b]{0.3\textwidth}
        \centering
        \includegraphics[width=\textwidth]{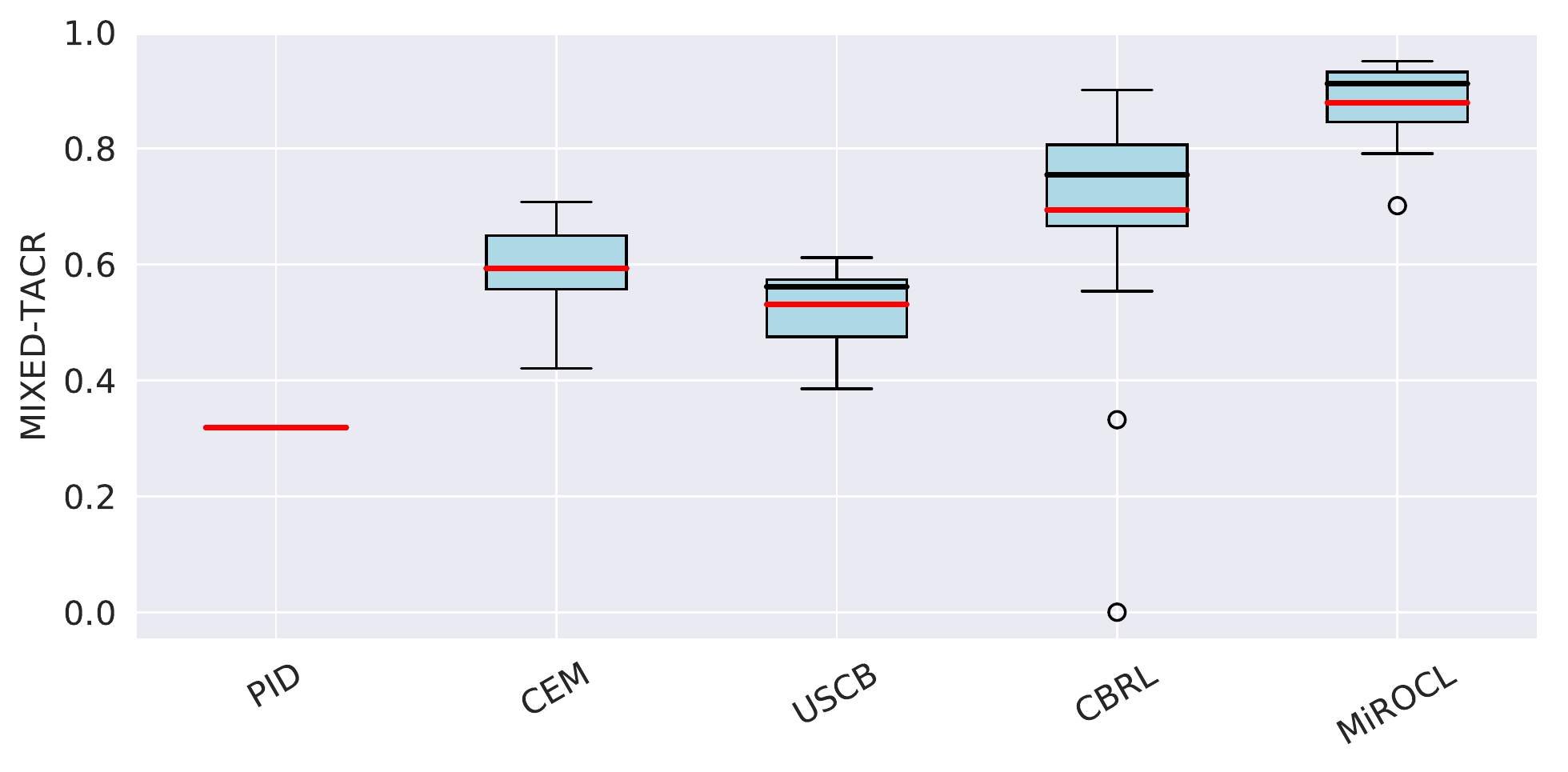}
    %  \caption{the conditional revenue improvement over oracle ANDR\wang{MCB needs update}}
    %  \label{fig:y equals x}
    \end{subfigure}
    \vspace{-.3cm}
    \caption{\small The results of CR@2\% (Left), IID-TACR/GSP-TACR (Middle), and OOD-TACR/MIX-TACR (Right) in the \ind{} (Top) and \syn{} (Bottom) settings are shown above. Each boxplot shows the average (red) and median (black) results of 20 independent repeated runs.}
    \label{fig:plots}
    \vspace{-.3cm}
\end{figure*}

}

\dataplot{}
\sotaplot{}
\synplot{}
\plots{}

\section{Derivations}
\subsection{Derivation for the teacher's objective}\label{app:teacher}
To begin with, we aim to find a distribution of environments that maximizes the regret of the current policy, within the Wasserstein ball around the empirical distribution in the embedding space. For probability meassures $P_1$  and $P_2$ supported on the latent space $\W$, and the couplings $\Pi(P_1,P_2)$, the Wasserstain distance over the metric space $\W$ is defined as:
\begin{equation}
    W_\kappa (P_1,P_2) \defeq \underset{H\in \Pi(P_1,P_2)}{\sup}\E_H\lb \kappa(\omega_1,\omega_2)\rb
\end{equation}
which finds the minimum cost to morph from the distribution $P_1$ to $P_2$, with the cost function $\kappa(\cdot,\cdot)$. 

The distance capture the geometry of the space $\W$ with the cost function, and we assume that $\kappa(x,y)=\|x-y\|_2$, i.e., the distance between two distributions relates with the euclidean distance between the samples of the two distributions. We define a set $\mathcal{P}$ of distributions as the $\rho$-ball around the empirical distribution $\bar{P}$, from which we search for the adversarial environments:
\begin{equation}
    \mathcal{P} = \{P:W_\kappa\left(P,\bar{P}\right)\le\rho\}
\end{equation}

The teacher's objective amounts to the following:
\begin{equation}
\begin{aligned}
    &\sup_{P(\boldo)} \E_{\boldo} \lb \reg (\pi,\boldo)\rb \\
    &= \sup_{P(\boldo)}\{
        \E_{\boldo} \lb \reg (\pi,\boldo)\rb 
        - \lambda W_\kappa(P,\bar{P})
    \}
\end{aligned}
\end{equation}
with Lagrangian relaxation parameter $\lambda\ge 0$. 

Based on the assumption that $\reg(\pi,\boldo)$ and $\kappa(\cdot,\cdot)$ are continuous, we have the following dual re-formulation~\cite{advtrn}, 
\begin{equation}
\begin{aligned}
    &\sup_{P:W_\kappa\left(P,\bar{P}\right)\le\rho} \E_{\boldo} \lb \reg (\pi,\boldo)\rb \\
    &=\underset{\gamma\ge 0}{\inf}\left\{
    \lambda\rho + 
    \underset{\bar{P}(\boldo^\prime)}{\E}
    \lb 
    \sup_\boldo \reg(\pi,\boldo)-\lambda\kappa(\boldo,\boldo^\prime)
    \rb
    \right\}.
\end{aligned}
\end{equation}
where the dual variable $\lambda\ge 0$. Then we have
\begin{equation}
\begin{aligned}
    \sup_{P(\boldo)}\{
        \E_{\boldo} \lb \reg (\pi,\boldo)\rb 
        - \lambda W_\kappa(P,\bar{P}) \\
        = \underset{\bar{P}(\boldo^\prime)}{\E}
    \lb 
    \sup_\boldo \reg(\pi,\boldo)-\lambda\kappa(\boldo,\boldo^\prime)
    \rb.
\end{aligned}
\end{equation}

\subsection{Derivation for the variational bounds}\label{sec:regbound}
%%%%%%%%%%%%%%%%%%%%%%%%%
% KL(pi,xi) = \sum pi\log pi - \sum pi\log xi
% cross entropy = \sum pi \log xi
% Detailed derivations, 
We aim to show that the objective in Eq.~\eqref{policyobj} amounts to minimizing an upper bound on the regret. We first show how we derive the lower bound from the discrepancy between the expert and the policy in the KL divergence. 
\begin{equation}
\begin{aligned}
    &\D_{KL}\left(\xi,\pi\right) 
    = \E_{\xi(a|o_t,h_t)}\lb \log \frac{\xi(a|o_t,h_t)}{\pi(a|o_t,h_t)}\rb \\
\end{aligned}
\end{equation}
where
\begin{equation}
\begin{aligned}
&\log \frac{\xi(a|o_t,h_t)}{\pi(a|o_t,h_t)}
    = \int_\omega q(\omega|h_t) \log \frac{\pi(a|o_t,h_t)}{\xi(a|o_t,h_t)} d\omega\\
& = \int_\omega q(\omega|h_t)
\log\left( \frac{\xi(a|o_t,h_t,\omega)}{\pi(a|o_t,h_t)}\cdot\frac{p(\omega|h_t)}{p(\omega|h^\ast)}\right) d\omega\\
& = \E_{q(\omega|h_t)}\lb - \log \pi(a|o_t,h_t,\omega)\rb - \E_{q(\omega|h_t)}\lb - \log \xi(a|o_t,h_t,\omega)\rb \\
&+\D_{KL}\left(q(\omega|h_t)\|p(\omega|h^\ast)\right) 
- \D_{KL}\left(q(\omega|h_t)\|p(\omega|h_t) \right)
\end{aligned}
\end{equation}
We can derive the bound of Eq.~\eqref{emb} similarly as the above shows. 

It follows that the discrepancy can factorizes into a cross entropy term between the expert and the policy, a constant term on the entropy of the expert, and two KL terms between the posteriors. 
\begin{equation}
\begin{aligned}
&\min_\pi  \D_{KL}\left(\xi\|\pi\right)=\min_\pi \E_{\xi,q}\lb - \log \pi_\phi(a|o_t,h_t,\omega) \rb +\text{const}\\
&+\D_{KL}\left(q(\omega|h_t)\|p(\omega|h^\ast)\right) 
- \D_{KL}\left(q(\omega|h_t)\|p(\omega|h_t) \right)
\end{aligned}\label{lowreg}
\end{equation}
We note that the first KL term aims to align with the causal structure of the expert. The second KL term is empirically found to perform worse, so we learn with only the first KL term, resulting in the upper bound~\eqref{policyobj}.

Then we introduce a theorem from the imitation learning literature that states the relationship of regret and KL divergence-based expert-policy discrepancy~\cite{imi1}.
\begin{thm}
Under a given MDP $\MO$, if $\E_{d^{\xi^\ast}_{\boldo}}\lb \D_{KL}\left(\xi^\ast,\pi\right) \rb \le \epsilon $, we have that $\reg(\pi;\MO) \le 2\sqrt{2}H^2\sqrt{\epsilon}$.
\end{thm}
The theorem\footnote{The bound is under the infinite sample situation, and can be further bounded using classical learning theory.} states that the regret is bounded by the expert-policy discrepancy measure. Accordingly, minimizing the objective~\eqref{policyobj} is directly minimizing an upper bound on the regret, which can indicate regret bound guarantee.

\section{Experiments}\label{sec:expimpl}
\textbf{Dataset.}
Fig.~\ref{fig:dataplot} shows the distribution shift in the in-distribution and out-of-distribution split of \ind{} dataset. For \syn{} dataset, We simulate the high and low periods of market competitions and ROI as sinusoidal functions with different amplitudes and phases. The dynamic mixed auction is constructed by sampling inflection points and then interpolating between the points. 

\noindent\textbf{Implementations}
The policy is implemented as a BERT transformer, which consists of an encoder for $q()$ and a decoder for $\pi_\phi()$. The ground-truth posterior $p()$ shares the encoder with $q()$ but is bi-directional, i.e., without future masking. We adopt entropy-regularized RL objective, which minimizes the KL divergence between the policy and the Boltzmann distribution of a state-action value function. The state-action value also takes as input the inferred $\omega_t$, meanwhile learned with the reward returned by the off-distribution reward estimator $r_\theta()$. 
% \plots{}

\noindent\textbf{Additional Results}
The TACR result on \syn{} dataset is shown in Fig.~\ref{fig:syn_tacr}, and other metrics are shown in Fig.~\ref{fig:plots}. 

%%%%%%%%%%%%%%%%%%%%%%%%%%%%%%

\end{document}